\documentclass{article}

\PassOptionsToPackage{numbers, compress}{natbib}

\usepackage[final]{neurips_2024}

\usepackage{amsmath,amsfonts,bm}

\def\eqref#1{equation~\ref{#1}}

\def\1{\bm{1}}

\DeclareMathAlphabet{\mathsfit}{\encodingdefault}{\sfdefault}{m}{sl}
\SetMathAlphabet{\mathsfit}{bold}{\encodingdefault}{\sfdefault}{bx}{n}

\usepackage{color}
\definecolor{deeppink}{rgb}{0.62, 1.00, 0.84} 
\definecolor{deep_blue}{rgb}{0.00, 0.45, 0.72} 

\newcommand{\LS}[1]{\left({#1}\right)}
\newcommand{\LM}[1]{\left\{{#1}\right\}}
\newcommand{\LL}[1]{\left[{#1}\right]}

\newcommand{\C}[1]{\mathcal{#1}}
\newcommand{\R}{\mathbb{R}}

\newcommand{\B}[1]{\boldsymbol{#1}}

\newcommand{\OP}[1]{\mathtt{#1}}

\usepackage[utf8]{inputenc} 
\usepackage[T1]{fontenc}    
\usepackage{hyperref}       
\usepackage{url}            
\usepackage{booktabs}       
\usepackage{amsfonts}       
\usepackage{nicefrac}       
\usepackage{microtype}      

\usepackage{amssymb}
\usepackage{here}
\usepackage{enumerate}
\usepackage{caption}
\usepackage{array}
\usepackage{multirow}
\usepackage{wrapfig}
\usepackage{amscd}
\usepackage[pdftex]{graphicx}
\usepackage{threeparttable}
\usepackage{listings}
\usepackage{amsthm} 
\usepackage{cleveref}
\usepackage{mathtools}
\usepackage{amsthm}
\usepackage{algorithm}
\usepackage{algorithmicx}
\usepackage{algpseudocode}
\usepackage{subcaption}
\usepackage{booktabs} 
\usepackage{vtable} 
\usepackage{tabularx}
\usepackage{pifont}
\usepackage[table]{xcolor}

\newtheoremstyle{PropositionNum}
    {\topsep}{\topsep}              
    {\itshape}                      
    {}                              
    {\bfseries}                     
    {.}                             
    { }                             
    {\thmname{#1}\thmnote{ \bfseries #3}}
\theoremstyle{PropositionNum}

\definecolor{deepskyblue}{RGB}{54, 125, 189}
\definecolor{lightskyblue}{RGB}{58, 178, 198}
\hypersetup{colorlinks = true, linkcolor = deepskyblue,
            urlcolor  = lightskyblue,
            citecolor = deepskyblue,
            anchorcolor = deepskyblue}

\title{RETR: Multi-View Radar Detection Transformer \\for Indoor Perception}

\author{%
    Ryoma Yataka$^{1,3,}$\thanks{Equal contribution.}\;\,$^{,}$\thanks{The work was done as a visiting scientist from ITC in Mitsubishi Electric Corporation.}\;\;,\;
    Adriano Cardace$^{2, \ast, }$\thanks{The work was done during his internship at MERL.}\;\;,
    Pu (Perry) Wang$^{1, \ast}$,\\
    \textbf{Petros Boufounos}$^{1}$\textbf{,}\;
    \textbf{Ryuhei Takahashi}$^{3}$\\
    $^1$Mitsubishi Electric Research Laboratories (MERL), USA\\
    $^2$Department of Computer Science and Engineering, University of Bologna, Italy\\
    $^3$Information Technology R\&D Center (ITC), Mitsubishi Electric Corporation, Japan
}

\begin{document}

\maketitle

\begin{abstract}
Indoor radar perception has seen rising interest due to affordable costs driven by emerging automotive imaging radar developments and the benefits of reduced privacy concerns and reliability under hazardous conditions (e.g., fire and smoke). However, existing radar perception pipelines fail to account for distinctive characteristics of the multi-view radar setting. In this paper, we propose Radar dEtection TRansformer (\textbf{RETR}), an extension of the popular DETR architecture, tailored for multi-view radar perception. RETR inherits the advantages of DETR, eliminating the need for hand-crafted components for object detection and segmentation in the image plane. More importantly, RETR incorporates carefully designed modifications such as 1) depth-prioritized feature similarity via a tunable positional encoding (TPE); 2) a tri-plane loss from both radar and camera coordinates; and 3) a learnable radar-to-camera transformation via reparameterization, to account for the unique multi-view radar setting. Evaluated on two indoor radar perception datasets, our approach outperforms existing state-of-the-art methods by a margin of $15.38+$ AP for object detection and $11.91+$ IoU for instance segmentation, respectively. 
Our implementation is available at \url{https://github.com/merlresearch/radar-detection-transformer}.
\end{abstract}

\section{Introduction}
\label{sec:introduction}

Perception information encompasses the processes and technologies to detect, interpret, and understand their surroundings.  Complementary to the mainstream camera and LiDAR sensors, radar can enhance the safety and resilience of perception under low light, adversarial weather (e.g., rain, snow, dust), and hazardous conditions (e.g., smoke, fire) at affordable device and maintenance cost. 
An emerging application of radar perception is indoor sensing and monitoring for elderly care, building energy management, and indoor navigation~\cite{Gurbuz2019_IndoorMonitoring}. A notable limitation of indoor radar perception is the low semantic features from radar signals. 

Earlier efforts use radar detection points~\cite{Mingmin2017_RFSleep, Sengupta2020_mmPose} to support simple classification tasks such as fall detection and activity recognition over a limited number of patterns. To support challenging perception tasks such as object detection, pose estimation, and segmentation, lower-level radar signal representation such as radar heatmaps is more preferred. Along this line, the earliest work is RF-Pose~\cite{Zhao2018_RFPose} using a convolution-based autoencoder network to fuse features from the two radar views and regress keypoints for 2D image-plane pose estimation. It is later extended to 3D human pose estimation~\cite{Zhao2018_RFPose3D}. It is noted that RF-Pose is not publicly accessible. More recently, RFMask~\cite{Wu2023_RFMask} borrows the Faster R-CNN framework~\cite{Ren2017_FasterRCNN} by proposing candidate regions only in the horizontal radar heatmap via a region proposal network (RPN). A corresponding proposal in the vertical radar heatmap is automatically determined using a \emph{fixed-height} candidate region at the same depth as the horizontal proposal. The combined horizontal and vertical proposals are then projected into the image plane for bounding box (BBox) estimation. In addition, RFMask calculates the BBox loss only over the 2D horizontal radar view and disregards features from the vertical radar heatmap for BBox estimation. 

\begin{figure*}[t]
    \centering
    \includegraphics[width=\textwidth]{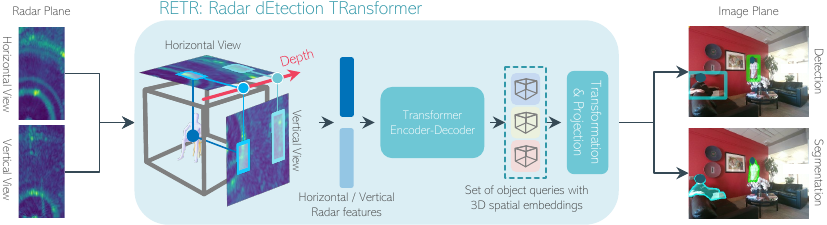}
    \caption{
    By taking horizontal-view and vertical-view radar heatmaps as inputs, RETR introduces a depth-prioritizing positional encoding (exploit the shared depth between the two radar views) into transformer self-attention and cross-attention modules and outputs a set of 3D-embedding object queries to support image-plane object detection and segmentation via a calibrated or learnable radar-to-camera coordinate transformation and 3D-to-2D pinhole camera projection. 
    } 
    \label{fig:concept}
\end{figure*}

In this paper, we exploit features from both horizontal and vertical radar views for object estimation and segmentation and introduce \textbf{Radar dEtection TRansformer (RETR)} (Fig.~\ref{fig:concept}). RETR extends the popular Detection Transformer (DETR)~\cite{Carion2020_detr}, which effectively eliminates the need for hand-crafted components such as non-maximum suppression and proposal/anchor generation, to the multi-view radar perception. More importantly, RETR incorporates carefully designed modifications to exploit the unique multi-view radar setting such as shared depth dimension and the transformation between the radar and camera coordinate systems. Our contributions are summarized below:
\begin{enumerate}
    \item 
    \textbf{Extending DETR for Multi-View Radar Perception:} 1) Encoder: we associate features from both radar views by applying self-attention over the pooled multi-view radar features, eliminating the need for a cumbersome association scheme. We introduce a top-$K$ feature selection to allow only $K$ features from each view to keep the complexity low. 2) Decoder: the DETR decoder provides a natural way to associate the same object query to corresponding features from the two radar views via cross-attention. As such, the object query is able to learn 3D spatial embedding of objects in the radar coordinate (see Fig.~\ref{fig:concept}). 
    \item \textbf{Tunable Positional Encoding}: To enhance feature association across the two radar views, we further exploit the fact that the two radar views share the depth dimension and introduce a tunable positional encoding (TPE) as an inductive bias. TPE imposes constraints in the attention map to prioritize the relative importance of depth dimension and avoid exhaustive correlations between radar views. 
    \item \textbf{Tri-Plane Loss from Both 3D Radar Coordinate and 2D Image Plane}: we enforce the output queries of the DETR decoder to directly predict 3D BBoxes in the radar coordinate system and convert them into the 2D image plane. We introduce a tri-plane loss that combines the BBox loss in the 3D radar plane and that in the 2D image plane, to calculate the global set-prediction loss. 
    \item \textbf{Learnable Radar-to-Camera Coordinate Transformation}: We employ a calibrated radar-to-camera coordinate transformation via a calibration process and a learnable coordinate transformation via reparameterization by preserving the orthonormal (i.e., 3D special orthogonal group $\C{SO}\LS{3}$) structure of the rotation matrix.
\end{enumerate}
We demonstrate the effectiveness of our contributions through evaluations on two open datasets: the HIBER dataset~\cite{Wu2023_RFMask} and the MMVR dataset~\cite{Perry2024_MMVR}.

\section{Related Work}
\label{sec:related_works}
\paragraph{Radar-based Object Detection and Segmentation:}
Indoor radar perception tasks include object detection (BBoxes), pose estimation (keypoints), and instance segmentation (human masks)~\cite{Fadel2015_RFCapture15,Vandersmissen2018_IndoorIdentification,Zhao2018_RFPose,Zhao2018_RFPose3D,Lu20, Ouaknine2021_MVRSS, Zhao21, Wu2023_RFMask, CubeLearn}, and radar datasets in different data formats were reported in~\cite{Vandersmissen2018_IndoorIdentification,Singh2019_RadHAR,Sengupta2020_mmPose,Xue2021_mmMesh,An2022_mRI,Yang2023_mmFi,Lee2023_HuPR, Wu2023_RFMask, Perry2024_MMVR}. Particularly, radar heatmap-based approaches have gained attention not only in indoor perception~\cite{Zhao2018_RFPose,Zhao2018_RFPose3D,Lee2023_HuPR,Wu2023_RFMask,Perry2024_MMVR} but also for  automotive radar perception~\cite{Liu2023_Echoes,Peak2023_KRadar,Skog2024_humandetection4dradar,Kong2024_RTNH,Ding2024_radaroccrobust3doccupancy}, due to richer semantic features compared to those extracted from sparse radar point clouds~\cite{Singh2019_RadHAR,Sengupta2020_mmPose,Xue2021_mmMesh,An2022_mRI, Li2022_TempoRadar, Yang2023_mmFi, Yataka2024_SIRA}.
RF-Pose~\cite{Zhao2018_RFPose} predicts human poses on the image plane using a convolution autoencoder-based architecture.
With the HIBER  dataset~\cite{Wu2023_RFMask}, RFMask considers proposal-based object detection and instance segmentation.  
More recently, MMVR \cite{Perry2024_MMVR} has been openly released to accelerate advancements in indoor radar perception.

\begin{figure*}[t]
    \centering
    \includegraphics[width=1.0\textwidth]{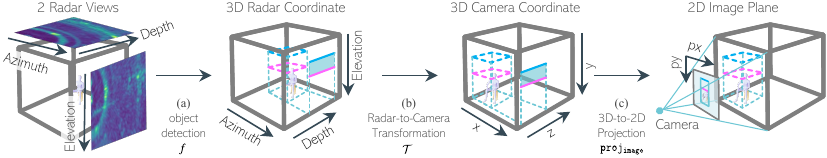}
    \caption{Indoor radar perception pipeline:  (a) multi-radar views are utilized to estimate 3D BBoxes in the radar coordinate system; (b) the 3D BBoxes are then transformed into the 3D camera coordinate system by a radar-to-camera transformation; and (c) the transformed 3D BBoxes are projected onto the image plane for final object detection. \textbf{\textcolor{deep_blue}{Blue line}} denotes a fixed-height regional proposal in RFMask, while \textbf{\textcolor{magenta}{Magenta line}} denotes an object query with learnble height in RETR.}
    \label{fig:concept_3d}
\end{figure*}

\paragraph{Image-based Object Detection and Segmentation with DETR:}
Since the introduction of DETR for 2D image-plane object detection, subsequent studies have been developed based on its framework~\cite{Meng2021_CondDETR,Zhu2021_DeformableDETR,Caron2021_DINO,Liu2022_DabDETR,Wang2022_anchorDETR,Liu2023_StableDETR,Hu2023_DacDETR,Pu2023_RankDETR}, largely due to DETR’s ability to eliminate the need for hand-designed components such as non-maximum suppression (NMS).
In~\cite{Meng2021_CondDETR}, Conditional DETR decomposes the roles of content and positional embeddings in the transformer decoder, improving not only prediction accuracy but also training convergence speed. More recently, ~\cite{Pu2023_RankDETR} has proposed Rank-DETR as a rank-oriented architectural design, guaranteeing lower false positives and false negatives in prediction.

\section{Preliminary}
\label{sec:preliminary}

\paragraph{Generation of Radar Heatmaps:}
Conceptually, let us consider a pair of (virtual) horizontal and vertical antenna arrays with $N_{\OP{ant}}$ elements for each array, sending a set of frequency modulated continuous waveform (FMCW) pulses for object detection~\cite{Perry2024_MMVR,Wu2023_RFMask,Sun2020_MIMORadar_AutonomousDriging}.
The two 1D arrays generate one horizontal radar view in the azimuth-depth $(x-z)$ domain and one vertical radar view in the elevation-depth $(y-z)$ domain,
\begin{equation}
y_{\OP{hor}}\LS{t, x, z} = \sum_{k=1}^{K_p}{\sum_{m=1}^{M}{s_{k,m,t} e^{j 2 \pi \frac{d_m\LS{x, z}}{\lambda_k}}}}, \quad
y_{\OP{ver}}\LS{t, y, z} = \sum_{k=1}^{K_p}{\sum_{m=1}^{M}{s_{k,m,t} e^{j 2 \pi \frac{d_m\LS{y, z}}{\lambda_k}}}},
\end{equation}
where $s_{k, m, t}$ denotes the $k$-th sample of FMCW sweep on the $m$-th antenna at time $t$, $\lambda_{k}$ is the wavelength of the $k$-th sample, $d_m\LS{x, z}$ denotes the round-trip distance from the $m$-th array element to a position $\LS{x, z}$, and $K_p$ and $M$ denote the number of samples and the number of array antennas, respectively. Usually, the azimuth $x$ is in an interval of $x \in \mathcal{X}= [x_{\text{min}}: \Delta x: x_{\text{max}}]$ and  the elevation $y$ and the depth $z$ are similarly defined. At a particular time $t$,  we have the horizontal radar heatmap $\B{Y}_{\OP{hor}}(t) = \{| y_{\OP{hor}}\LS{t, x, z}| \}_{x\in {\cal{X}}}^{z\in {\cal{Z}}} \in {\cal{R}}^{W\times D}$ and the vertical radar heatmap $\B{Y}_{\OP{ver}}(t) = \{| y_{\OP{ver}}\LS{t, y, z}| \}_{y\in {\cal{Y}}}^{z\in {\cal{Z}}} \in {\cal{R}}^{H\times D}$ with a shared depth axis. The multi-view radar testbeds in HIBER \cite{Wu2023_RFMask} and MMVR \cite{Perry2024_MMVR} utilize advanced MIMO-FMCW radar systems. We defer the MIMO-FMCW radar heatmap generation to Appendix~\ref{sec:process_mvr}.

\paragraph{Indoor Radar Perception:} By taking $T$ consecutive multi-view radar heatmaps ($\B{Y}_{\OP{hor}}\in {\cal{R}}^{T\times W\times D}$ and $\B{Y}_{\OP{ver}}\in {\cal{R}}^{T\times H\times D}$) as the input, we are interested in detecting objects on the image plane:
\begin{equation}
    \B{F}_{\OP{image}} = \OP{proj}_{\OP{image}}\LS{\C{T}\LS{f\LS{\B{Y}_{\OP{hor}}, \B{Y}_{\OP{ver}}}}},
\end{equation}
where $\B{F}_{\OP{image}}$ denotes predicted BBoxes for object detection and pixel-level masks for instance segmentation.  Using the BBox as an example in Fig.~\ref{fig:concept_3d}, our pipeline includes the following steps: 1) Fig.~\ref{fig:concept_3d} (a): By taking the two radar views over $T$ consecutive frames $\LS{\B{Y}_{\OP{hor}}, \B{Y}_{\OP{ver}}}$ as input, the end-to-end object detection module $f$ outputs a set of parameters describing 3D BBoxes in the radar coordinate system; 2) Fig.~\ref{fig:concept_3d} (b): The radar-to-camera 3D coordinate transformation $\C{T}$ converts the predicted 3D BBoxes at the output of $f$ to corresponding 3D BBoxes in the 3D camera coordinate system; 3) Fig.~\ref{fig:concept_3d} (c): The 3D-to-2D projection $\OP{proj}_{\OP{image}}$ projects the 3D BBox in the camera coordinate system into corresponding 2D image plane normally with a known pinhole camera model. 

\section{RETR: Radar Detection Transformer}
\label{sec:proposed_method}
We first present the RETR architecture and then highlight radar-oriented modifications. We defer the discussion on Segmentation to Appendix~\ref{sec:seg}. 

\subsection{RETR Architecture}
\label{sec:multi_view_transformer}
\begin{figure*}[t]
    \centering
    \includegraphics[width=1.0\textwidth]{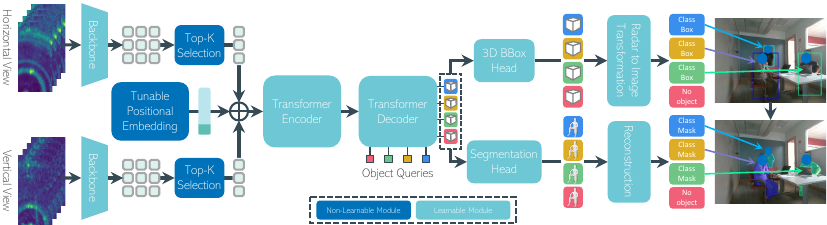}
    \caption{The RETR architecture: 1) \textbf{Encoder}: Top-$K$ features selection and tunable positional encoding to assist feature association across the two radar views; 2) \textbf{Decoder}:  TPE is also used to assist the association between object queries and multi-view radar features;  3) \textbf{3D BBox Head}: Object queries are enforced to estimate 3D objects in the radar coordinate and projected to $3$ planes for supervision via a coordinate transformation; 4) \textbf{Segmentation Head}: The same queries are used to predict binary pixels within each predicted BBox in the image plane.} 
    \label{fig:concept_arch}
\end{figure*}

We present the RETR architecture in Fig.~\ref{fig:concept_arch}, introducing its major modules in a left-to-right order. Refer to Appendix~\ref{sec:discussion_retr} for the detailed architecture. 

\paragraph{Backbone:} Given $\B{Y}_{\OP{hor}}\in {\cal{R}}^{T\times W\times D}$ and $\B{Y}_{\OP{ver}}\in {\cal{R}}^{T\times H\times D}$, a shared  backbone network (e.g., ResNet~\cite{He2016_ResNet}) generates separate horizontal-view and vertical-view radar feature maps:
$\B{Z}_{\OP{hor}} = \OP{backbone}\LS{\B{Y}_{\OP{hor}}}\in\R^{C\times \frac{W}{s}\times \frac{D}{s}}$ and $\B{Z}_{\OP{ver}} = \OP{backbone}\LS{\B{Y}_{\OP{ver}}}\in \R^{C\times \frac{H}{s}\times \frac{D}{s}}$, where $C$ and $s$ represent the number of channels and downsampling ratio over the spatial dimension, respectively. 

\paragraph{Feature Selection:} A transformer-based encoder expects a sequence of features as input. This is done by mapping the feature maps into a sequence of $P$ \emph{multi-view radar features} $\B{H} = \LM{\B{H}_{\OP{hor}}, \B{H}_{\OP{ver}}} \in \R^{C\times P}$: 
$\B{Z}_{\OP{hor}} \rightarrow \B{H}_{\OP{hor}} \in \R^{C\times P_{\OP{hor}}}$ and $\B{Z}_{\OP{ver}} \rightarrow \B{H}_{\OP{ver}}\in \R^{C\times P_{\OP{ver}}}$, where $P=P_{\OP{hor}}+P_{\OP{ver}}$.  We defer the discussion of top-$K$ feature selection to Section~\ref{sec:topK}. 

\paragraph{Encoder as Cross-View Radar Feature Association:} The transformer encoder provides a simple yet effective method for associating radar features from both horizontal and vertical views by applying self-attention over the pool of $P$ multi-view radar features $\B{H}= \LM{\B{H}_{\OP{hor}}, \B{H}_{\OP{ver}}} \in \R^{C\times P}$, eliminating the need for cumbersome association schemes. Specifically, the $l$-th ($l=0,\cdots,L_{\OP{self}}-1$) encoder layer updates the multi-view radar features through multi-head self-attention $\OP{Att}_{\OP{self}}$:
\begin{equation}
    \label{eq:enc_self_att}
    \B{H}^{l+1} = \bar{\B{H}}^l + \OP{FFN}\LS{\bar{\B{H}}^l},\quad \bar{\B{H}}^l = \B{H}^l+\OP{Att}_{\OP{self}}\LS{\OP{Que}\LS{\B{H}^l}, \OP{Key}\LS{\B{H}^l}, \OP{Val}\LS{\B{H}^l}},
\end{equation}
where $\OP{FFN}$ denotes feed-forward networks, $L_{\OP{self}}$ is the number of encoder layers, and $\OP{Que}$, $\OP{Key}$ and $\OP{Val}$ are projections to derive the multi-head query, key and value embedding from $\B{H}$, respectively. For the first ($0$-th) layer, we have $\B{H}^0= \B{H}$. Note that we omit the description of ``Layer norm'' and ``multi-head index'' in Eq.~\ref{eq:enc_self_att} for clarity. 

Additionally, since the multi-view radar features lack positional information and the self-attention is permutation-invariant, we supplement $\B{H}^{l}$ with positional embedding added (or attached) to the input of each encoder layer. Refer to Section~\ref{sec:pos_embed} for a tunable positional encoding.

\paragraph{Decoder to Associate Object Queries with Multi-View Radar Features:} The decoder provides a natural way to associate the same object query with features from the two radar views via cross-attention. For each decoder layer, it takes $N$ object queries $\B{Q}^{l} = \LM{\B{q}_{1}, \cdots, \B{q}_{N}}\in\R^{C\times N}$ as its input, and consists of a self-attention layer, a cross-attention layer and a FFN. Specifically for the $l$-th ($l=0,1,\cdots,L_{\OP{cross}}-1$) decoder layer, it first updates all queries through multi-head self-attention:
\begin{equation}
    \label{eq:dec_self_att}
    \bar{\B{Q}}^l = \B{Q}^l+\OP{Att}_{\OP{self}}\LS{\OP{Que}\LS{\B{Q}^l}, \OP{Key}\LS{\B{Q}^l}, \OP{Val}\LS{\B{Q}^l}},
\end{equation}
where $\OP{Que}$, $\OP{Key}$ and $\OP{Val}$ are the projections with different parameterization from those in the self-attention layer (Eq.~\ref{eq:enc_self_att}).
Then, the decoder layer further updates the object queries $\bar{\B{Q}}^l$ of Eq.~\ref{eq:dec_self_att} via multi-head cross-attention with the multi-view radar features $\B{H}^{L_{\OP{self}}}$ from the encoder output:
\begin{equation}
\label{eq:dec_cross_att}
    \B{Q}^{l+1} = \tilde{\B{Q}}^l + \OP{FFN}\LS{\tilde{\B{Q}}^l},\quad \tilde{\B{Q}}^l = \bar{\B{Q}}^l+\OP{Att}_{\OP{cross}}\LS{\OP{Que}\LS{\bar{\B{Q}}^l}, \OP{Key}\LS{\B{H}^{L_{\OP{self}}}}, \OP{Val}\LS{\B{H}^{L_{\OP{self}}}}},
\end{equation}
where both $\bar{\B{Q}}^l$ and $\B{H}^{L_{\OP{self}}}$ are supplemented with positional embedding. 
Finally, the decoder outputs $N$ enhanced object queries $\B{Q}^{L_{\OP{cross}}}$ for downstream tasks. 

\paragraph{Mapping from 3D Radar Coordinate to 2D Image Plane:} Given the $N$ enhanced object queries $\B{Q}^{L_{\OP{cross}}}$, RETR directly estimates 3D BBoxes in the radar coordinate:
\begin{equation}
    \bar{\B{g}} = \LM{cx, cy, cz, w, h, d}^{\top} = \OP{sigmoid}\LS{\OP{FFN}\LS{\B{q}}}, \quad \B{q} \in \B{Q}^{L_{\OP{cross}}}
\end{equation}
where $\bar{\B{g}} $ describes the 3D BBox center and respective widths along the 3D axes, and $\OP{sigmoid}$ normalizes the 3D BBox prediction to $\LL{0,1}$. 
Then, as shown in Fig.~\ref{fig:concept_3d} (b), we apply a radar-to-camera transformation $\C{T}$ to convert the predicted 3D BBoxes to ones in the 3D camera coordinate as
\begin{equation} \label{c2r}
    \B{g}_{\OP{camera}}^{i} = \LM{x^i_{\OP{camera}}, y^i_{\OP{camera}}, z^i_{\OP{camera}}}^{\top} = \C{T}\LS{\B{g}_{\OP{radar}}^{i}} = \B{R}\B{g}_{\OP{radar}}^{i} + \B{t}, \quad i=1,2, \cdots, 8,
\end{equation}
where $\B{R}$ is a 3D rotation matrix, $\B{t}\in\R^3$ is the 3D translation vector, and $\B{g}_{\OP{radar}}^{i}$ is $i$-th corner of the 3D BBox corresponding to $\bar{\B{g}}$. Subsequently in Fig.~\ref{fig:concept_3d} (c),  we project the 3D BBoxes $\B{g}_{\OP{camera}}^{i}$ onto the 2D image plane via a 3D-to-2D projection. 
From the projected 2D corners, one can calculate the 2D BBox center and width and height in the image plane as 
\begin{align}
\B{b}_{\OP{init}} = \LM{cx, cy, w, h}^{\top} =\OP{proj}_{\OP{image}}\LS{\B{G}_{\OP{camera}}}. 
\end{align}
The final BBox estimation $\widehat{\B{b}}_{\OP{image}}$ in the image plane is obtained by adding an offset head $\OP{FFN}:\R^{10}\to\R^{4}$ to compensate for the spatial downsampling and normalizing it to the interval $\LL{0,1}$:
\begin{equation}
\label{bhat}
\widehat{\B{b}}_{\OP{image}} = \OP{sigmoid}\LS{\B{b}_{\OP{init}} + \OP{FFN}\LS{\B{b}_{\OP{init}}\oplus\bar{\B{g}}}}.
\end{equation}

\subsection{Top-$K$ Feature Selection}
\label{sec:topK}
In DETR, the sequentialization simply collapses the spatial dimensions of the feature map into a single dimension, resulting in $P_{\OP{hor}}=WD/s^2$ and $P_{\OP{ver}}=HD/s^2$ features for the horizontal and vertical radar feature maps, respectively. As a result, we have $P=(W+H)D/s^2$ multi-view radar features. It is known that the complexity of transformers grows quadratically with respect to the feature length $P$.  
Here,  we introduce a customized Top-$K$ feature selection, maintaining a low complexity for the RETR encoder and decoder:
$\B{H}_{\OP{hor}}=\OP{Selector}\LS{\B{Z}_{\OP{hor}}}\in\R^{C\times K}$ and $\B{H}_{\OP{ver}}=\OP{Selector}\LS{\B{Z}_{\OP{ver}}}\in\R^{C\times K}$, where $K \ll \min \{ WD/s^2, HD/s^2\}$. In this case, we shrink the multi-view radar tokens from $P=(W+H)D/s^2$ to $P=2K$. 
For each radar frame, we consistently select the Top-$K$ strongest features, which may originate from varying spatial locations depending on the specific radar frame. Consequently, the gradient propagates back through the selected $K$ features to the backbone weights, irrelevant to their spatial locations.

\subsection{TPE: Tunable Positional Encoding}
\label{sec:pos_embed}
\begin{figure*}[t]
    \centering
    \includegraphics[width=\textwidth]{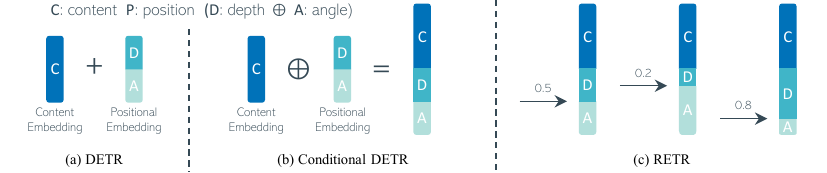}
    \caption{Schemes of positional encoding: (a) the sum operation in the original DETR; (b) the concatenation in Conditional DETR; and (c) TPE in RETR that allows for adjustable dimensions between depth and angular embeddings and promotes higher similarity scores for keys and queries with similar depth embeddings than those far apart in depth.}
    \label{fig:concept_tpe}
\end{figure*}

The TPE is built on the top of the concatenation operation between the content embedding $\B{c}$ (either feature embedding $\B{h}$ at the encoder or 
decoder embedding 
$\B{q}$ at the decoder) and positional embedding $\B{p}$ in the conditional DETR~\cite{Meng2021_CondDETR} (see Fig.~\ref{fig:concept_tpe} (b)): 
\begin{equation} \label{condPos}
    \LS{\B{c}_{\OP{que}} \oplus \B{p}_{\OP{que}}}^{\top} \LS{\B{c}_{\OP{key}} \oplus \B{p}_{\OP{key}}} 
    = \B{c}_{\OP{que}}^{\top}\B{c}_{\OP{key}} + \B{p}_{\OP{que}}^{\top}\B{p}_{\OP{key}},
\end{equation}
where $\oplus$ denotes concatenation, rather than the sum in DETR~\cite{Carion2020_detr} (see Fig.~\ref{fig:concept_tpe} (a)):
\begin{equation} \label{detrPos}
    \LS{\B{c}_{\OP{que}} + \B{p}_{\OP{que}}}^{\top}\LS{\B{c}_{\OP{key}} + \B{p}_{\OP{key}}} 
    = \B{c}_{\OP{que}}^{\top}\B{c}_{\OP{key}} + \B{c}_{\OP{que}}^{\top}\B{p}_{\OP{key}} + \B{p}_{\OP{que}}^{\top}\B{c}_{\OP{key}} + \B{p}_{\OP{que}}^{\top}\B{p}_{\OP{key}}.
\end{equation}
It is seen that Eq.~\ref{condPos} eliminates the cross terms between the content and positional embeddings in Eq.~\ref{detrPos} and, allowing content/positional embeddings focus on their respective attention weights, contributes to faster training convergence~\cite{Meng2021_CondDETR}. 

In our case, the positional embedding is composed of a depth ($y$) axis and an angular (either azimuth $x$ or elevation $z$) axis. As such, $\B{p} = \B{d} \oplus \B{a}$ 
with $\B{d}$ representing the depth positional embedding and $\B{a}$ the angular positional embedding. Then expanding Eq.~\ref{condPos} with $\B{p} = \B{d} \oplus \B{a}$  leads to 
\begin{equation} \label{tpe}
    \LS{\B{c}_{\OP{que}} \oplus \B{d}_{\OP{que}} \oplus \B{a}_{\OP{que}}}^{\top} \LS{\B{c}_{\OP{key}} \oplus \B{d}_{\OP{key}} \oplus \B{a}_{\OP{key}}} 
    = \B{c}_{\OP{que}}^{\top}\B{c}_{\OP{key}} + \B{d}_{\OP{que}}^{\top}\B{d}_{\OP{key}} + \B{a}_{\OP{que}}^{\top}\B{a}_{\OP{key}}. 
\end{equation}

In Eq.~\ref{tpe}, we have the following observations: 
\begin{enumerate}
\item $ \B{c}_{\OP{que}}^{\top}\B{c}_{\OP{key}}$ reflects how similar the features in the key and query may appear; 
\item Depth similarity $\B{d}_{\OP{que}}^{\top}\B{d}_{\OP{key}}$ remains consistent regardless of whether the key and query originate from the same radar view or different radar views; 
\item Angular similarity $\B{a}_{\OP{que}}^{\top}\B{a}_{\OP{key}}$ can be a self-angular similarity (azimuth-to-azimuth or elevation-to-elevation) when the key and query are from the same radar view, or a cross-angular similarity (azimuth-to-elevation or elevation-to-azimuth) for different radar views.
\end{enumerate}
Motivated by the above observations, we can promote higher similarity scores for keys and queries with similar depth embeddings than those far apart in depth, especially for the ones from different views, by allowing for adjustable dimensions between depth and angular embeddings: 
\begin{align}
d_{\OP{dep}} = \alpha d_{\OP{pos}}, \quad  d_{\OP{ang}}=(1-\alpha) d_{\OP{pos}} \quad \rightarrow \quad d_{\OP{dep}}  + d_{\OP{ang}}= d_{\OP{pos}},
\end{align}
where the tunable dimension ratio $\alpha$ is in the interval $[0, 1]$. As illustrated in Fig.~\ref{fig:concept_tpe} (c), when $\alpha=0.5$, the positional embedding is equivalent to that used in conditional DETR. When $\alpha$ approaches $0$, the depth positional embedding is minimized, making the depth similarity $\B{d}_{\OP{que}}^{\top}\B{d}_{\OP{key}}$ negligible in Eq.~\ref{tpe}. Conversely, as $\alpha$ approaches $1$, the depth positional embedding dimension increases, and so does the importance of the depth similarity in Eq.~\ref{tpe}. 

We implement our TPE with a fixed sine/cosine positional encoding along the depth and angular (azimuth or elevation) dimension. For an even depth/angular positional dimension, we have
\begin{align}
    \B{d}_{2i} & = \sin({\OP{p_{dep}}}/{\tau^{{2i}/{d_{\OP{dep}}}}}), \quad \B{d}_{2i+1} = \cos({\OP{p_{dep}}}/{\tau^{{2i}/{d_{\OP{dep}}}}}), \quad i=0,1, \cdots,  d_{\OP{dep}}/2-1, \label{depPos} \\
    \B{a}_{2i} & = \sin({{\OP{p_{\OP{ang}}}}/{\tau^{{2i}/{d_{\OP{ang}}}}}}), \quad \B{a}_{2i+1} = \cos({{\OP{p_{\OP{ang}}}}/{\tau^{{2i}/{d_{\OP{ang}}}}}}), \quad i=0,1, \cdots,  d_{\OP{ang}}/2-1, \label{angPos}
\end{align}
where $\OP{p}_{\OP{dep}/\OP{ang}}$ and $d_{\OP{dep}/\OP{ang}}$ are the position index and dimension for the depth and angular axes, respectively, $i$ is the (even/odd) element index, and $\tau$ = 10000 is a temperature. By adjusting the ratio $\alpha$ in Eq.~\ref{tpe}, we change the dimensions of the depth $\B{d}$ in Eq.~\ref{depPos} and the angular $\B{a}$ in Eq.~\ref{angPos}, while keeping the total positional dimension of $\B{p}=\B{d} \oplus \B{a}$ constant. We show the visualization of TPE in Appendix~\ref{sec:tpe}.

\subsection{Tri-Plane Set-Prediction Loss}
\label{sec:loss}

DETR calculates a matching cost matrix with each element constructed from 1) a classification cost $\C{L}_{\OP{class}}$ and 2) a BBox loss between one of $N$ predictions $\widehat{\B{b}}$ and one of ground truth BBoxes $\B{b}$ (including the ``no object'' class). The BBox loss is a weighted combination of the generalized intersection over union (GIoU) loss $\C{L}_{\OP{GIoU}}$~\cite{Rezatofighi2019_giou} and the $\ell_1$ loss $\C{L}_{\OP{L_1}}$:
\begin{align} \label{bbox}
    \C{L}_{\OP{box}}(\B{b}, \widehat{\B{b}}) = \lambda_{\OP{GIoU}} \C{L}_{\OP{GIoU}}(\B{b}, \widehat{\B{b}}) + \lambda_{\OP{L_1}} \C{L}_{\OP{L_1}}(\B{b}, \widehat{\B{b}}),
\end{align}
where $\lambda_{*}$ denotes the weight. Over the permutation set $\mathfrak{S}_{N}$ between $N$ predictions and ground truth objects, the Hungarian algorithm~\cite{kuhn1955_hungarian} is applied with the matching cost matrix to find the optimal assignment $\sigma^* \in \mathfrak{S}_{N}$ of predictions to ground truth. 
Given $\sigma^*$ , the loss is computed only for the matched pairs and is referred to as the set-prediction loss. 

\begin{wrapfigure}[11]{r}{2.4in}
    \vspace{-0.2in}
    \centering
    \includegraphics[width=0.44\textwidth]{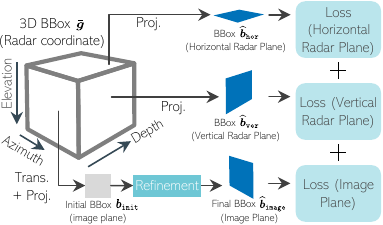}
    \caption{Tri-Plane BBox loss.}
    \label{fig:triplane_loss}
\end{wrapfigure}
Since RETR predicts 3D BBoxes $\bar{\B{g}}$ in the 3D radar coordinate and maps them into the 2D image plane, we propose to enhance the above Hungarian match cost matrix using a \emph{Tri-Plane BBox Loss} from both the radar coordinate and image plane. 
This is illustrated in Fig.~\ref{fig:triplane_loss}, where a 3D BBox $\bar{\B{g}}$ in the radar coordinate is projected onto 1) the 2D horizontal radar plane as $\widehat{\B{b}}_{\OP{hor}} = \OP{proj}_{\OP{hor}}(\bar{\B{g}})$ (the top branch); 2) the 2D vertical radar plane as $\widehat{\B{b}}_{\OP{ver}} = \OP{proj}_{\OP{ver}}(\bar{\B{g}})$ (the middle branch); and 3) the 2D image plane as $\widehat{\B{b}}_{\OP{image}}$ of Eq.~\ref{bhat} (the bottom branch). The tri-plane BBox loss $\C{L}_{\OP{box}}^{\OP{tri}}$ sums up 2D BBox losses over all three planes using Eq.~\ref{bbox}:
\begin{equation}
    \label{eq:loss}
    \C{L}_{\OP{box}}^{\OP{tri}} = 
    \C{L}_{\OP{box}}\LS{\B{b}_{\OP{hor}}, \widehat{\B{b}}_{\OP{hor}}} + \C{L}_{\OP{box}}\LS{\B{b}_{\OP{ver}}, \widehat{\B{b}}_{\OP{ver}}} + \C{L}_{\OP{box}}\LS{\B{b}_{\OP{image}}, \widehat{\B{b}}_{\OP{image}}}.
\end{equation}
RETR finds the optimal assignment $\sigma^*_{\OP{tri}}$ using the matching cost with 1) the original classification cost $\C{L}_{\OP{class}}$ and 2) the tri-plane BBox loss  $\C{L}_{\OP{box}}^{\OP{tri}}$. The resulting set-prediction loss using $\sigma^*_{\OP{tri}}$ is referred to as the tri-plane set-prediction loss. 

\subsection{Learnable Radar-to-Camera Coordinate Transformation}
\label{sec:learnTrans}
The rotation matrix $\B{R}$ and translation vector $\B{t}$ in the radar-to-camera transformation of Eq.~\ref{c2r} can be calibrated in advance.
However, this calibration process may be accurate only for a limited interval of depth and angles. Instead of relying on the calibrated transformation, we introduce a learnable transformation via a reparameterization on $\B{R}$ while keeping it orthonormal.
To this end, we need to ensure that the learnable $\widehat{\B{R}}$ resides in the 3D special orthogonal group $\C{SO}\LS{3}$. 
Considering that $\C{SO}\LS{3}$ is a special case of a Lie group, one of the differentiable manifolds, {we can firstly map a 3D vector $\B{\omega}=\LM{\omega_x, \omega_y, \omega_z}^\top\in\R^3$ to Lie algebra $\mathfrak{so}\LS{3}$ using the projection $\LL{\cdot}:\R^3\to\mathfrak{so}\LS{3}$. And then we apply the exponential map $\OP{exp}:\mathfrak{so}\LS{3}\to\C{SO}\LS{3}$ that maps $[\B{\omega}]$ into the nearest point in $\C{SO}\LS{3}$ such that the resulting $\OP{exp} \LS{\LL{\B{\omega}}}$ resides on $\C{SO}\LS{3}$} and satisfies the orthonormal structure~\cite{Lee2003_manifold,sola2021_microlietheorystate}. This leads to the following reparameterization of $\widehat{\B{R}}$ in terms of $\B{\omega}$:
\begin{equation}
   \widehat{\B{R}} \approx \OP{exp} \LS{\LL{\B{\omega}}}=\B{I}+\frac{\sin \phi}{\phi}\LL{\B{\omega}}+\frac{1-\cos \phi}{\phi^2}\LL{\B{\omega}}^2,\;\text{s.t.}\; \LL{\B{\omega}}=\left[\begin{array}{ccc}
    0 & -\omega_z & \omega_y \\
    \omega_z & 0 & -\omega_x \\
    -\omega_y & \omega_x & 0
    \end{array}\right], 
\end{equation}
where 
$\phi=\|\B{\omega}\|$ is the $\ell_2$ norm, With the above reparameterization, the learnable radar-to-camera coordinate transformation in Eq.~\ref{c2r} reduces to learn the vector $\B{\omega}$ and the translation vector $\B{t}$. 

\section{Experiments}
\label{sec:experiments}

\subsection{Setup}
\label{sec:setting}

\paragraph{Datasets:}
We evaluate performance over two open indoor radar perception datasets: MMVR\footnote{\url{https://zenodo.org/records/12611978}}~\cite{Perry2024_MMVR} and HIBER\footnote{\url{https://github.com/Intelligent-Perception-Lab/HIBER}}~\cite{Wu2023_RFMask}. MMVR includes multi-view radar heatmaps collected from over $20$ human subjects across $6$ rooms over a span of $9$ days. In our implementation, we utilize data from \textbf{Protocol 2} (P2) which includes $237.9$K data frames capturing both single and multiple human subjects in diverse activities such as walking, sitting, stretching, and writing on the board. For the training-validation-test split, we follow the data split \textbf{S1} as defined in MMVR.

HIBER, partially released, includes multi-view radar heatmaps from $10$ human subjects in a single room but from different angles with two data splits: 1) ``WALK'', consisting of $73.5K$ data frames with one subject (Section~\ref{sec:main_results}); and 2) ``MULTI'', consisting of $70.8K$ radar frames with multiple ($2$) human subjects walking in the room (Appendix~\ref{sec:additional_ablation_study}). 
More dataset details can be found in Appendix~\ref{sec:experimental_setting}. 

\paragraph{Implementation:}
We consider RFMask~\cite{Wu2023_RFMask} and DETR~\cite{Carion2020_detr} as baseline methods. Since RFMask and DETR originally compute the BBox loss only in the 2D horizontal ($\OP{H}$) radar plane and the 2D image ($\OP{I}$) plane, respectively, we enhance both  methods with a unified bi-plane BBox loss ($\OP{H}+\OP{I}$). We also introduce a DETR variant with top-$K$ feature selection, allowing it to take features from both horizontal ($\OP{H}$) and vertical ($\OP{V}$) heatmaps as input. For RETR, we set $K=256$ for the top-$K$ selection, the positional embedding dimension to $d_{\OP{pos}}=256$, and a tunable dimension ratio at $\alpha=0.6$. We include one variant that only employs the TPE at the decoder (TPE@Dec.). More hyper-parameter settings can be found in Appendix~\ref{sec:experimental_setting}.

\paragraph{Metrics:} For object detection, we adopt average precision (AP) at two IoU thresholds of $0.5$ ($\OP{AP_{50}}$) and $0.75$ ($\OP{AP_{75}}$) and its mean ($\OP{AP}$) over thresholds $[0.5: 0.05:0.95]$. We also consider average recall (AR) when it is restricted to making only one detection  ($\OP{AR_1}$) or up to $10$ detections ($\OP{AR_{10}}$) per image. For segmentation, we report the average $\OP{IoU}$ value between the predictive and ground truth masks. Detailed metric definitions can be found in Appendix~\ref{sec:definition_metrics}.

\subsection{Main Results}
\label{sec:main_results}

\begin{table}[t]
    \footnotesize
    \centering
    \caption{Main results of object detection in the image plane under ``P2S1'' of MMVR. The top section shows results from conventional models, while the bottom section presents RETR results. }
    \setlength\tabcolsep{8.2pt}
    \begin{tabular}{lccc| ccc cccc}
        \toprule
        \multicolumn{1}{c}{$\OP{Model}$} & $\OP{Dim}$ & $\OP{Input}$ & $\OP{BBox} \ \OP{Loss}$ & $\OP{AP}$ & $\OP{AP_{50}}$ & $\OP{AP_{75}}$ & $\OP{AR_{1}}$ & $\OP{AR_{10}}$ \\ 
        \midrule
        RFMask & 2D & $\OP{H}$, $\OP{V}$ & $\OP{H}+\OP{I}$ & 31.37 & 61.50 & 27.48 & 33.23 & 38.41 \\ 
        DETR & 2D & $\OP{H}$ & $\OP{H}+\OP{I}$ & 29.38 & 62.31 & 25.35 & 31.32 & 43.06 \\ 
        DETR (Top-$K$) & 2D & $\OP{H}$, $\OP{V}$ & $\OP{H}+\OP{I}$ & 39.71 & 82.74 & 33.29 & 38.98 & 52.81 \\ 
        \midrule
        RETR (TPE@Dec.) &3D & $\OP{H}$, $\OP{V}$ & $\OP{H}+\OP{V}+\OP{I}$ & 45.94 & 81.99 & 44.04 & 42.03 & 57.38 \\ 
        RETR & 3D & $\OP{H}$, $\OP{V}$ & $\OP{H}+\OP{V}+\OP{I}$ & \cellcolor{gray!20}46.75 & \cellcolor{gray!20}83.80 & \cellcolor{gray!20}46.06 & \cellcolor{gray!20}42.19 & \cellcolor{gray!20}57.39 \\
        \bottomrule
    \end{tabular}
    \label{tab:main_mmvr_p2s1}
\end{table}

\begin{table}[t]
    \footnotesize
    \centering
    \caption{Main results of object detection in the image plane  under ``WALK'' of HIBER. The notation follows the same format as Table~\ref{tab:main_mmvr_p2s1}. }
    \setlength\tabcolsep{8.pt}
    \begin{tabular}{lccc| ccc cccc c}
        \toprule
        \multicolumn{1}{c}{$\OP{Model}$} & $\OP{Dim}$ & $\OP{Input}$ & $\OP{BBox}$ \ $\OP{Loss}$ & $\OP{AP}$ & $\OP{AP_{50}}$ & $\OP{AP_{75}}$ & $\OP{AR_{1}}$ & $\OP{AR_{10}}$ \\
        \midrule
        RFMask & 2D & $\OP{H}$, $\OP{V}$ & $\OP{H}+\OP{I}$& 17.77 & 52.46 & 6.78 & 32.71 & 32.71 \\
        DETR & 2D & $\OP{H}$ & $\OP{H}+\OP{I}$ & 14.45 & 47.33 & 4.25 & 28.64 & 28.64 \\
        DETR (Top-$K$) & 2D & $\OP{H}$, $\OP{V}$ & $\OP{H}+\OP{I}$ & 14.35 & 48.94 & 5.50 & 28.78 & 28.78 \\
        \midrule
        RETR (TPE@Dec.) & 3D & $\OP{H}$, $\OP{V}$ & $\OP{H}+\OP{V}+\OP{I}$ & 20.18 & 52.53 & 7.32 & 32.91 & 32.91 \\
        RETR & 3D & $\OP{H}$, $\OP{V}$ & $\OP{H}+\OP{V}+\OP{I}$ & \cellcolor{gray!20}22.09 & \cellcolor{gray!20}59.83 & \cellcolor{gray!20}10.99 & \cellcolor{gray!20}35.16 & \cellcolor{gray!20}35.16 \\
        \bottomrule
    \end{tabular}
    \label{tab:main_hiber_walk}
\end{table}

\begin{figure*}[t]
    \centering
    \includegraphics[width=\textwidth]{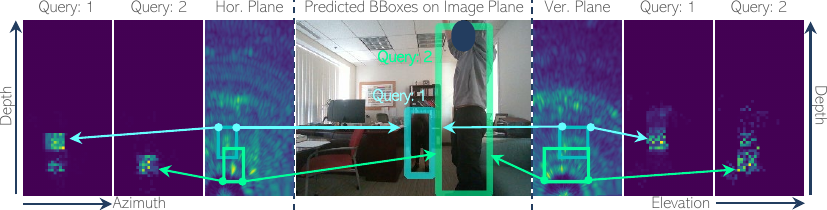}
    \caption{
    Visualization of cross-attention map between predicted BBoxes and multi-view radar features. BBoxes with the same color correspond to the same subject.
    } 
    \label{fig:cross_attentin}
\end{figure*}

\begin{table*}[t]
    \footnotesize
    \subfloat[
        A dimension \textbf{ratio} $\alpha=0.6$ of TPE achieves the best.
    \label{tab:ex_abl_ratio}
    ]{
    \setlength\tabcolsep{4pt}
    \begin{minipage}{0.4\linewidth}{\begin{center}
        \centering
        \begin{tabular}{ccccc}
        \toprule
        
        Ratio $\alpha$ & $0.2$ & $0.4$ & $0.6$ & $0.8$\\
    
        \midrule
        $\OP{AP}$ & 44.35 & 45.16 & \cellcolor{gray!20}46.75 & 43.81\\
        $\OP{AR1}$ & 42.04 & 41.60 & \cellcolor{gray!20}42.19 & 41.89\\
        \bottomrule
        \end{tabular}
        \end{center}}
    \end{minipage}
    }
    \hspace{0.5em}
    \subfloat[
        \textbf{Learnable Transformation} ($\OP{LT}$) may replace calibration.
    \label{tab:abl_mmvr_learnable_transformation}
    ]{
    \setlength\tabcolsep{8pt}
    \begin{minipage}{0.25\linewidth}{
        \centering
        \begin{tabular}{ccc}
        \toprule
        
        \textbf{$\OP{LT}$} & $\OP{AP}$ & $\OP{AR_1}$ \\
    
        \midrule
        - & 42.97 & 40.20  \\
        \checkmark & \cellcolor{gray!20}46.75 & \cellcolor{gray!20}42.19\\
        \bottomrule
        \end{tabular}}
    \end{minipage}
    }
    \hspace{0.5em}
    \subfloat[
        \textbf{Tri-plane loss} enhances object detection.
    \label{tab:abl_3d_prediction}
    ]{
    \setlength\tabcolsep{3pt}
    \begin{minipage}{0.26\linewidth}{\begin{center}
        \centering
        \begin{tabular}{ccc}
        \toprule
        
        $\OP{Loss}$ & $\OP{AP}$ & $\OP{AR_{1}}$ \\
    
        \midrule
        $\OP{H}+\OP{I}$ & 42.18 & 39.39 \\
        $\OP{H}+\OP{V}+\OP{I}$ & \cellcolor{gray!20}46.75 & \cellcolor{gray!20}42.19\\
        \bottomrule
        \end{tabular}
        \end{center}}
    \end{minipage}
    }
    \caption{\textbf{Ablation studies} under ``P2S1'' on MMVR.}
\end{table*}

\paragraph{MMVR:}
Table~\ref{tab:main_mmvr_p2s1} shows the main results on the MMVR dataset under ``P2S1''. 
Compared with RFMask, DETR with a single horizontal radar view does not show performance improvement. By just adding the vertical radar view at the input, DETR with top-$K$ selection exhibits a noticeable performance improvement over RFMask. Built upon DETR (Top-K), RETR (TPE@Dec.) implements two enhancements: 1) TPE at the decoder and 2) tri-plane BBox loss, resulting in further improvements with a gain of $6.23$ in $\OP{AP}$, $10.75$ in $\OP{AP_{75}}$, and $4.57$ in $\OP{AR_{10}}$, highlighting the importance of TPE and supervision at the vertical radar view. By further incorporating TPE at the encoder, the full version of RETR achieves an impressive performance improvement over RFMask, demonstrating increases of $15.38$ in $\OP{AP}$, $22.30$ in $\OP{AP_{50}}$, and $18.58$ in $\OP{AP_{75}}$, respectively. 
The results under ``P2S2'' on MMVR can be seen in Appendix~\ref{sec:additional_ablation_study}.

\paragraph{HIBER:}
Table~\ref{tab:main_hiber_walk} presents the main results on the HIBER dataset under ``WALK''. Similar to Table~\ref{tab:main_mmvr_p2s1}, we observe a similar trend of performance improvement from DETR to RETR variants. Numerically, we see increases of $4.32$ in $\OP{AP}$, $7.37$ in $\OP{AP_{50}}$, and $4.21$ in $\OP{AP_{75}}$, when directly comparing RETR to RFMask. These performance improvements are smaller compared with those in Table~\ref{tab:main_mmvr_p2s1}. This is potentially because the HIBER data under ``WALK'' predominantly involves walking, where RFMask's fixed-height vertical proposals may work fine. In contrast, MMVR under ``P2'' includes more diverse activities such as sitting, leading to likely overestimated vertical proposals for RFMask and thus greater improvements in MMVR than HIBER. 
The results under ``MULTI'' on HIBER can be seen in Appendix~\ref{sec:additional_ablation_study}.

\paragraph{Visualization of Cross-Attention Map:}
 
Fig.~\ref{fig:cross_attentin} presents the cross-attention map at the last decoder layer between predicted BBoxes (via object queries) and multi-view radar features. 
RETR accurately predicts the subject in the background of the image plane (middle panel) with a forward-bending posture (Query 1). 
The cross-attention maps of Query 1, with respect to horizontal (left) and vertical (right) radar features,  highlight areas with features contributing the most to Query 1. These contributing areas in the vertical plane are more stretched along the depth axis compared with those in the horizontal plane. Notably, the contributing areas from the two views share similar depth intervals. 
For Query 2 which identifies the subject in the foreground, the cross-attention maps shift its focus to contributing areas at closer depth compared with those for Query 1, indicating an effective 3D spatial embedding of object queries at the RETR output. We provide more visualizations in Appendix~\ref{sec:visualization_result}.

\paragraph{Limitation:} We present failure cases in Fig.~\ref{fig:vis_results_for_detection_fail} of Appendix~\ref{sec:visualization_result}. Predicting arm positions remains challenging, suggesting that RETR may not focus its attention  on regions with weak radar reflections. 
Moreover, multi-path reflections from the ground, ceiling, and other strong scatterers (e.g., metal) can cause (first-order or second-order) ghost targets and elevate the noise floor. Traditional signal processing techniques can mitigate these effects but require access to raw radar data. Alternatively, ghost targets can be labeled in the multi-view radar heatmaps, though this can be time-consuming and costly.  One can then extend RETR to classify output queries to one of $\{\emptyset, person, ghost\}$, alongside regressing queries to the BBox parameters.

\subsection{Ablation Studies}
\label{sec:ablation_study}

We report ablation studies with RETR under ``P2S1'' on MMVR.
Further results of ablation studies can be seen in Appendix~\ref{sec:additional_ablation_study}.

\paragraph{Tunable Dimension Ratio $\alpha$:}
Table~\ref{tab:ex_abl_ratio} presents the ablation study of the tunable dimension ratio $\alpha$ and its impact on the object detection performance in terms of $\OP{AP_{50}}$ ( primary vertical axis) and $\OP{AP}$ and $\OP{AP_{75}}$ (secondary vertical axis). The results indicate that $\alpha=0.6$ yields the best performance. The detection performance gradually decreases as $\alpha$ approaches to $0$ and $1$. 

\paragraph{Learnable Transformation ($\OP{LT}$):}
To evaluate the effectiveness of the Learnable Transformation in Section~\ref{sec:learnTrans}, we compare $\OP{AP}$ and $\OP{AR_1}$ metrics of RETRs with and without $\OP{LT}$. The results in Table~\ref{tab:abl_mmvr_learnable_transformation} indicate that it is possible to incorporate the radar-to-camera geometry into the end-to-end radar perception pipeline without the need for a cumbersome calibration step, while still achieving comparable perception performance.

\paragraph{Tri-Plane Loss for RETR:}
Table~\ref{tab:abl_3d_prediction} compares RETR with a bi-plane BBox loss (horizontal radar plane and image plane) to that with the tri-plane loss (including the vertical radar plane).  The results highlight the necessity of accounting for the vertical BBox loss and the importance of leveraging features from the vertical radar heatmap, leading to a performance improvement of $4.47$ in $\OP{AP}$.

\section{Conclusion}
\label{sec:conclusion}
In this paper, we  introduced RETR, extending DETR to the multi-view radar perception with carefully designed modifications such as depth-prioritized feature similarity via TPE, a tri-plane loss from  radar and camera coordinates, and a learnable radar-to-camera transformation. Experimental results over two radar datasets and comprehensive ablation studies demonstrate that RETR significantly outperforms both RFMask and DETR baseline methods. 

\paragraph{Broader Impacts:}
Indoor radar perception technologies, including RETR, offer a wide range of social applications in navigating and monitoring subjects such as the elderly, infants, robots, and humanoids, enhancing safety and energy efficiency while preserving privacy. However, it is crucial that perception results remain secure and private to prevent misuse in inferring subject attributes such as gender, size, and height. These technologies could potentially be used to advance indoor surveillance without individuals' acknowledgment or consent.

\bibliography{neurips}
\bibliographystyle{neurips}

\clearpage

\appendix


\section{Details of RETR Architecture}
\label{sec:discussion_retr}

\begin{figure*}[b]
    \centering
    \includegraphics[width=1.0\textwidth]{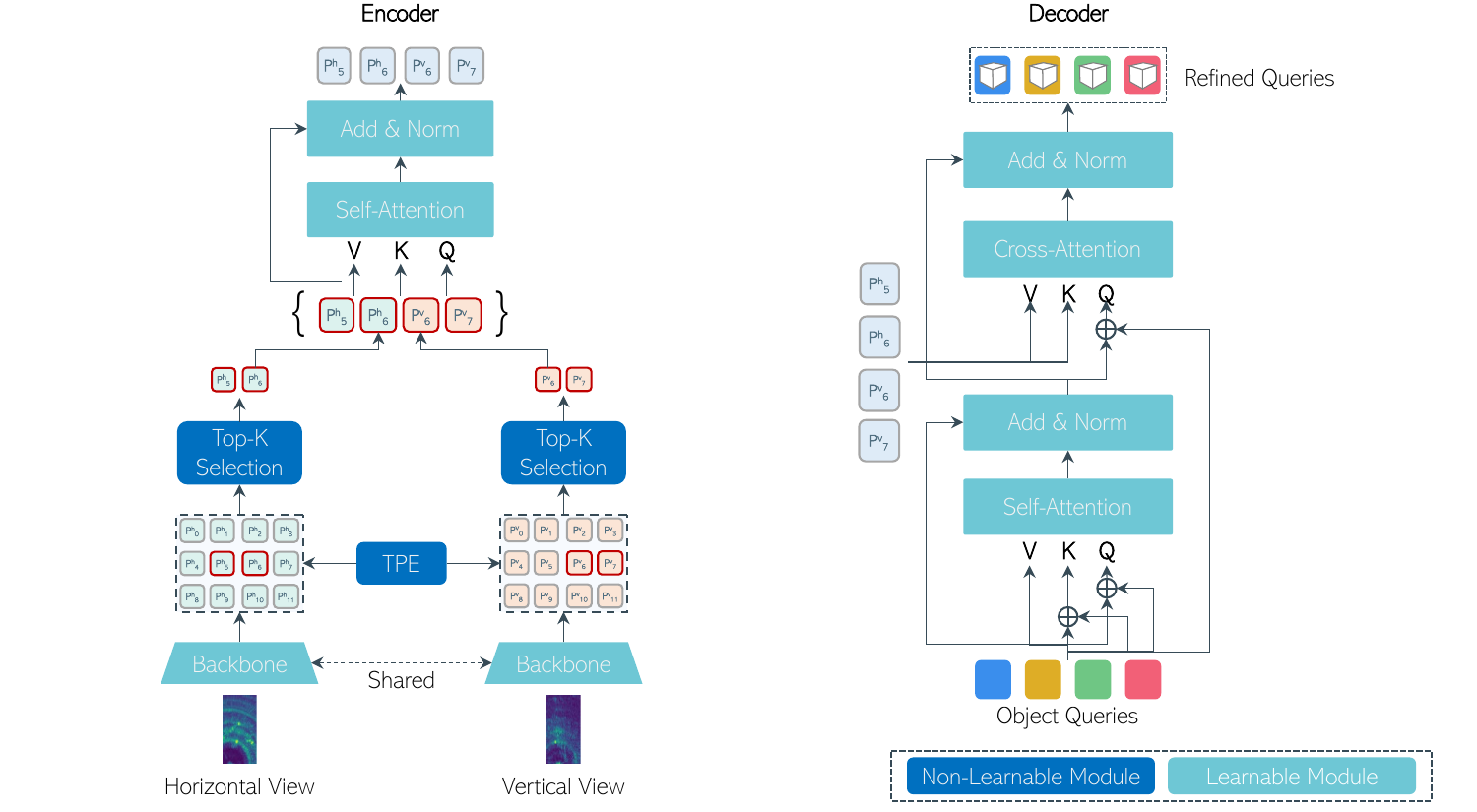}
    \caption{Illustration of (left) encoder and (right) decoder of RETR.} 
    \label{fig:arch_enc_dec}
\end{figure*}

\paragraph{Transformer Encoder and Decoder:}
Fig.~\ref{fig:arch_enc_dec} illustrates the transformer encoder and decoder used in RETR. 
In the original DETR implementation, the image features from the CNN backbone are given in input to the transformer encoder, with spatial positional embeddings added to the queries and keys at each multi-head self-attention layer of the encoder. 
On the other hand, RETR extracts features from a shared-weight backbone for both horizontal and vertical views and obtains them as $\{ P^{\OP{h}}_{1},\cdots,P^{\OP{h}}_{WD/s^2}, P^{\OP{h}}_{1},\cdots,P^{\OP{v}}_{HD/s^2} \}$. At this time, the positional encoding (TPE) is concatenated with the features (content). Subsequently, Top-$K$ selection is applied to extract the most relevant features and reduce time and space complexity (i.e., $P^{\OP{h}}_{5}, P^{\OP{h}}_{6}, P^{\OP{v}}_{6}$ and $P^{\OP{v}}_{7}$ in the left figure). These Top-$K$ features from the horizontal and vertical views are concatenated to compose a single sequence of tokens, which are then fed to the transformer encoder. The encoder consists of a stack of multi-head self-attention layers, that allow for the consideration of correlations between the two views.
The multi-head attention is simply the concatenation of $M$ single attention heads followed by a projection layer $L$ to regain the initial dimensionality. The common practice~\cite{Vaswani2017_attention} is to use residual connections, $\OP{dropout}$, and layer normalization:
\begin{align}   
    \OP{mhAtt} =& \OP{layernorm}\left(\OP{Que}\LS{\B{H}}+\OP{dropout}\left(L \tilde{\B{H}}\right)\right),\\
    \tilde{\B{H}} =& \OP{Att}\LS{\OP{Que}\LS{\B{H}}, \OP{Key}\LS{\B{H}}, \OP{Val}\LS{\B{H}}, \B{W}_1} \oplus \cdots \notag \\
    &\oplus \OP{Att}\LS{\OP{Que}\LS{\B{H}}, \OP{Key}\LS{\B{H}}, \OP{Val}\LS{\B{H}}, \B{W}_M},
\end{align}
where $\oplus$ is concatenation along the channel axis, and $\B{W}$ denotes the weight tensor of attention.

The decoder receives the decoder embeddings, which we initially set to zero and concatenated with the object queries, and encoder memory (i.e. the output sequence of the encoder transformer), generating refined embeddings through multiple multi-head self-attention and cross-attention layers.
In particular, the cross-attention layer utilizes the encoder memory to produce Keys and Values, which correlate with the Queries to produce the Refined Queries. In right figure of Fig.~\ref{fig:arch_enc_dec}, the decoder embeddings which concatenated with the object queries are first input into the self-attention, and the output is then passed through a normalization layer. At this point, the values are added using a residual structure. Next, cross-attention between the encoder memory, used as the key, and the decoder embeddings is calculated. Similarly, a residual structure is employed as in the self-attention. This entire sequence is repeated $L_{\OP{cross}}$ times to obtain the final decoder embeddings.

\paragraph{Computational Complexity}
Following the computational complexity notation used in the DETR paper, every self-attention mechanism in the encoder has a complexity of $\mathcal{O}(d^2 2K + d (2K)^2)$ where $d$ is the embedding dimension and $K$ is the number of selected features from the Top-$K$ selection. The cost of computing a single query/key/value embedding is $\mathcal{O}(d' d)$ (with $d=Md'$ where $M$ denotes the number of attention heads and $d'$ the dimension in each head), while the cost of computing the attention weights for one head is $\mathcal{O}(d' (2K)^2)$. Other computations may be negligible. In the decoder, each self-attention mechanism has a complexity of $\mathcal{O}(d^2 N + d N^2)$ where $N$ is the number of queries, and the cross-attention between query and multi-view radar features has a complexity of  $\mathcal{O}(d^2(N + 2K) + d 2NK)$. 
In conclusion, the overall complexity of our RETR model is 
\begin{equation}
    \mathcal{O}(4d^2 K + 4d K^2 + 2d^2 N + d N^2  + 2d NK).
\end{equation}

\section{Segmentation}
\label{sec:seg}

\begin{figure*}[t]
    \centering
    \includegraphics[width=1.0\textwidth]{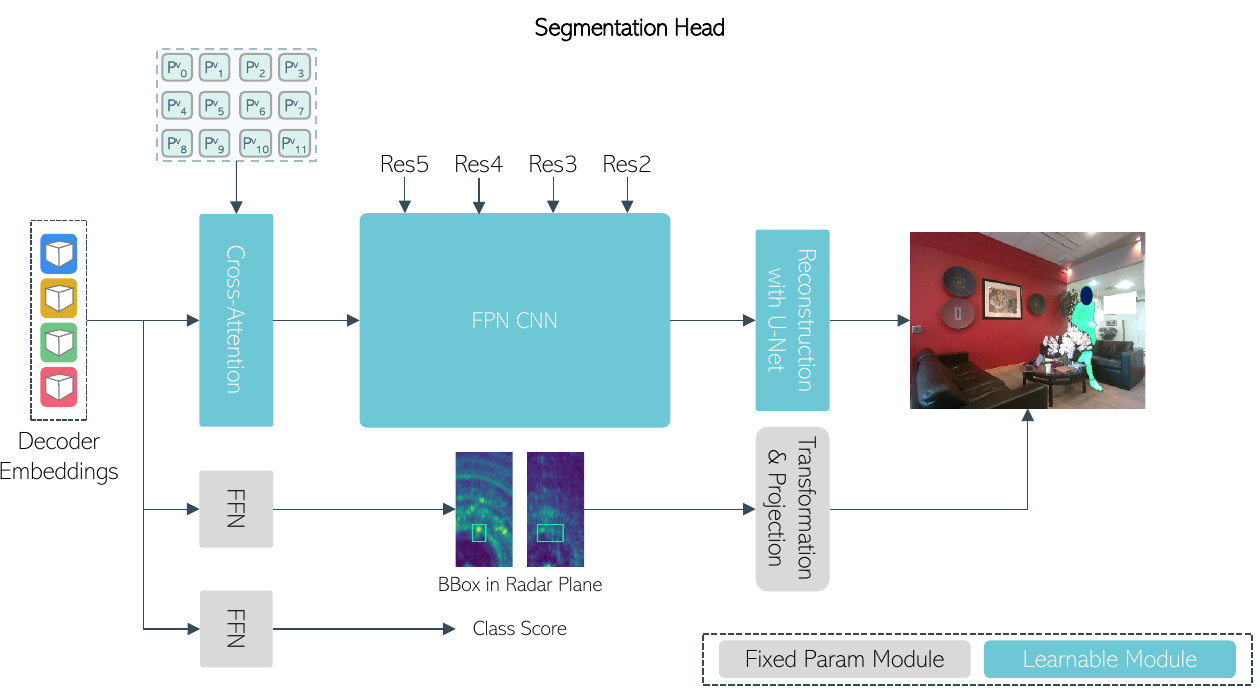}
    \caption{Illustration of segmentation head.} 
    \label{fig:arch_segmentation}
\end{figure*}

\paragraph{Architecture of Segmentation Head:} The original DETR is naturally extended by adding a segmentation head on top of the decoder outputs. Following this extension, our RETR enables segmentation by adding an architecture with a similar structure. Fig.~\ref{fig:arch_segmentation} illustrates the segmentation architecture we implemented, consisting of a cross-attention layer, a feature pyramid network (FPN)-style CNN, and final light U-Net~\cite{Ronneberger2015_unet}. Given a single refined query, we use a cross-attention layer to generate attention heatmaps for each object at a low resolution.
For the backbone output used in cross-attention, we utilized features extracted from the vertical heatmap, enhancing robustness to the height of the human. To increase the resolution of the mask, an FPN-style architecture is employed which also exploits the low-level backbone features at different layers (from 5 to 2) to generate some coarse segmentation masks. Since the FPN module is also responsible for lifting features from the radar view to the image plane, it does not have enough capacity to generate fine-grained segmentation masks. Thereby, we also add a very light U-Net to further refine the previously generated masks.
It is important to note that our model, differently from the original DETR implementation, predicts a single binary mask for each query.
Indeed, we exploit for each query the corresponding bounding box prediction in the radar plane, apply the Radar-to-Camera transformation and the 3D-to-2D image projection, to obtain the bounding box in the image plane.
This bounding box is finally used to extract the corresponding portion from the ground truth segmentation mask, which is employed to supervise the segmentation prediction for the same query. 
As a loss function, we adopt the DICE/F-1 loss~\cite{Milletari2016_vnet} and focal loss~\cite{Lin2017_focalLoss}.

\paragraph{Training:}
We note that the segmentation head can be trained at the same time as the BBox head in an end-to-end manner, or we can first train the detection head and then freeze all weights and train only the segmentation head in a two-step process. We followed the original DETR and employed the latter strategy.
During prediction, we filter out the detection with a confidence below 50\%, then compute the per-pixel argmax to produce the final binary segmentation mask.

\paragraph{Main Results:}

\begin{figure*}[t]
    \centering
    \includegraphics[width=1.0\textwidth]{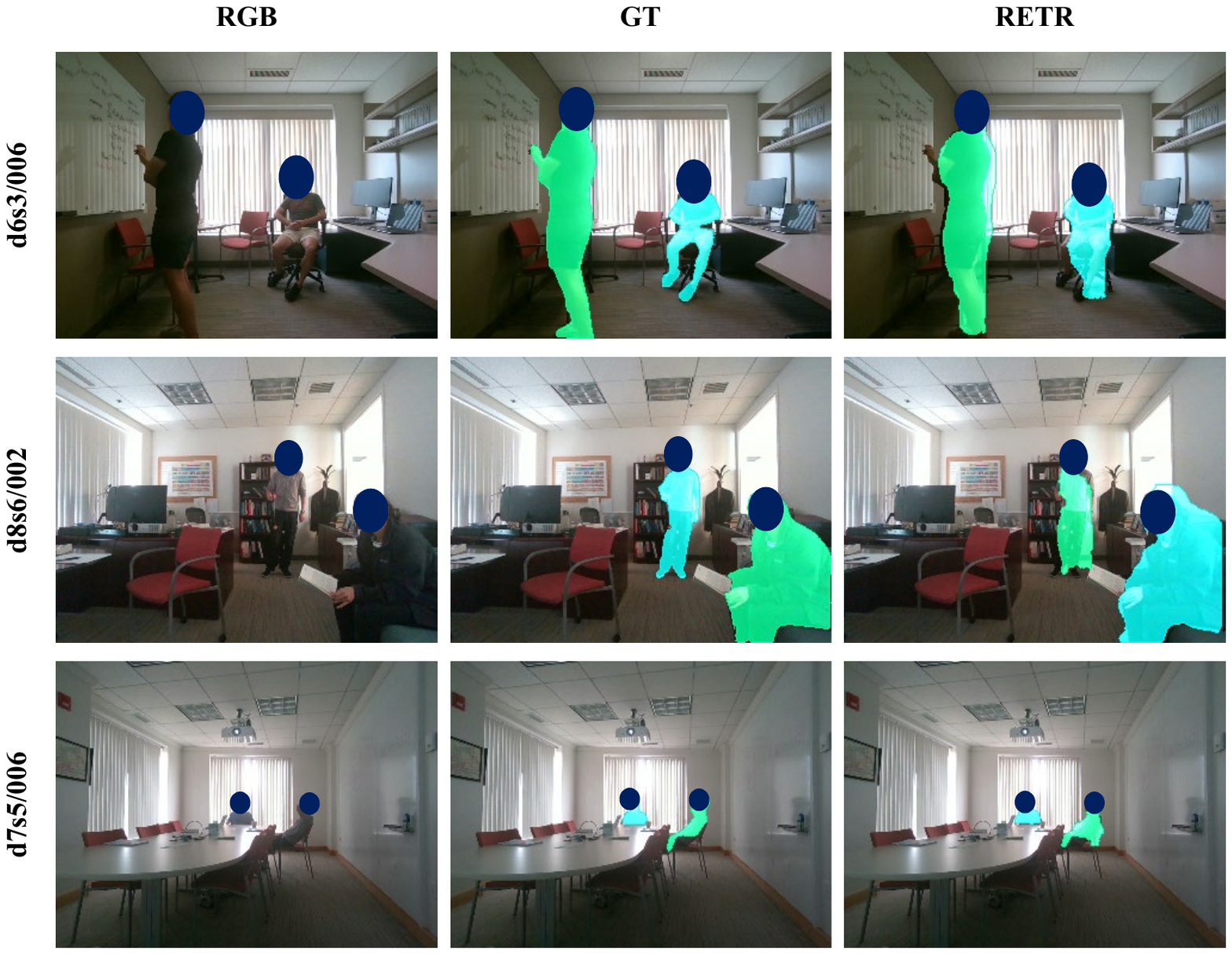}
    \caption{Visualization of segmentation results.} 
    \label{fig:vis_results_for_appendix_segmentation}
\end{figure*}

We report quantitative results the segmentation tasks in Table~\ref{tab:main_mmvr_p2s1_IoU}.
From this table, RETR (which combines all our contributions) achieves 77.07@$\OP{IoU}$, which is a significant performance improvement over the conventional RFmask with a gap of 11.77@$\OP{IoU}$. 
In addition, we point out how the DETR (Top-$K$) version (row 3) alone is able to increase the performance by almost 5\%.
We visualize the segmentation results in Fig.~\ref{fig:vis_results_for_appendix_segmentation}. 
Each row represents the data segment number in MMVR. It can be observed that RETR captures the shape of people with high fidelity. 
Notably, the results for d6s3 and d8s6 demonstrate that we are able to segment even complex postures, including sitting positions. Additionally, as shown in d7s5, RETR accurately estimates positions even when subjects are sitting far from the radar, such as at the back of the room. These results indicate that RETR can be easily extended from a detector to a segmentation model by adding a segmentation head, and it can accurately estimate masks.
For more visualizations, including comparison with RFMask and failure cases, see Appdendix~\ref{sec:visualization_result}.

\begin{table}[t]
    \footnotesize
    \centering
    \caption{Segmentation results under ``P2S1'' on MMVR.}
    \begin{tabular}{lccc| c}
        \toprule
        \multicolumn{1}{c}{$\OP{Model}$} & $\OP{Dim}$ & $\OP{Input}$ & $\OP{BBox} \ \OP{Loss}$ & $\OP{IoU}$ \\
        \midrule
        RFMask & 2D & $\OP{H}$, $\OP{V}$ & $\OP{H}+\OP{I}$ & 65.30 \\
        DETR & 2D & $\OP{H}$ & $\OP{H}+\OP{I}$ & 70.15 \\
        DETR (Top-$K$) & 2D & $\OP{H}$, $\OP{V}$ & $\OP{H}+\OP{I}$ & 75.76 \\
        \midrule
        RETR (TPE@Dec.) &3D & $\OP{H}$, $\OP{V}$ & $\OP{H}+\OP{V}+\OP{I}$ & 76.16 \\
        RETR & 3D & $\OP{H}$, $\OP{V}$ & $\OP{H}+\OP{V}+\OP{I}$ & \cellcolor{gray!20}77.21 \\
        \bottomrule
    \end{tabular}
    \label{tab:main_mmvr_p2s1_IoU}
\end{table}

\section{Visualization of TPE}
\label{sec:tpe}
\begin{figure*}[t]
    \centering
    \includegraphics[width=\textwidth]{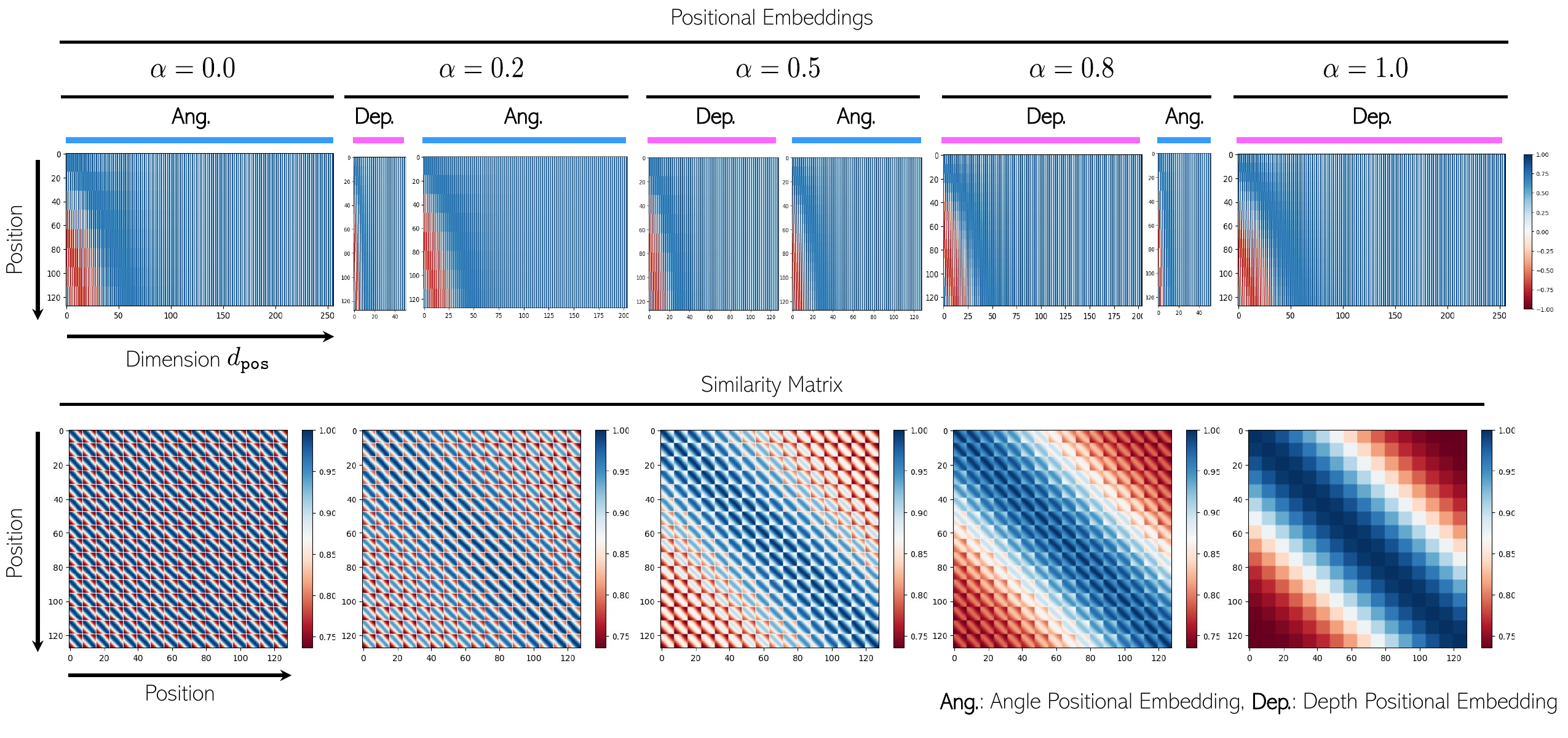}
    \caption{Visualization of TPE: (top row) positional embeddings for each axis; depth and angle, (bottom row) similarity through positions with dot product of positional embeddings. Each column denotes the $\alpha$ ($\alpha=0.0, 0.2, 0.5, 0.8, 1.0$). The blue color represents the large value, and red color represents the low value. The range is $\LL{-1, 1}$.}
    \label{fig:vis_tpe}
\end{figure*}

We visualize the positional embedding of each axis to observe the TPE. We calculated the positional embedding according to Eq.~\ref{depPos} and Eq.~\ref{angPos}, and visualized each axis as a separate figure with several value of $\alpha$ ($\alpha=0.0, 0.2, 0.5, 0.8, 1.0$). Fig.~\ref{fig:vis_tpe} shows the results. The top row is positional embeddings for each axis; depth and angle, and the bottom row is similarity through positions with dot product of positional embeddings. Each column denotes the $\alpha$ ($\alpha=0.0, 0.2, 0.5, 0.8, 1.0$). The blue color represents the large value, and red color represents the low value. 
The top row show that the characteristic elements are concentrated in the first some dimensions from top row. In addition, $\alpha=0.8$ and $\alpha=1.0$ are expected to contain more depth features since the spread of depth features is larger than $\alpha=0.5$ or lower.
Furthermore, when we look at the similarity matrix (bottom row), the deeper blue color is concentrated in the center of the matrix at $\alpha=0.8$ and $\alpha=1.0$. This indicates that the depths are more closely matched to each other, and that the degree of similarity can be changed by changing the $\alpha$.

\section{Multi-View MIMO-FMCW Radar Heatmap Generation}
\label{sec:process_mvr}
\begin{wrapfigure}[12]{r}{3.3in}
    \vspace{-0.14in}
    \centering
    \includegraphics[width=0.6\textwidth]{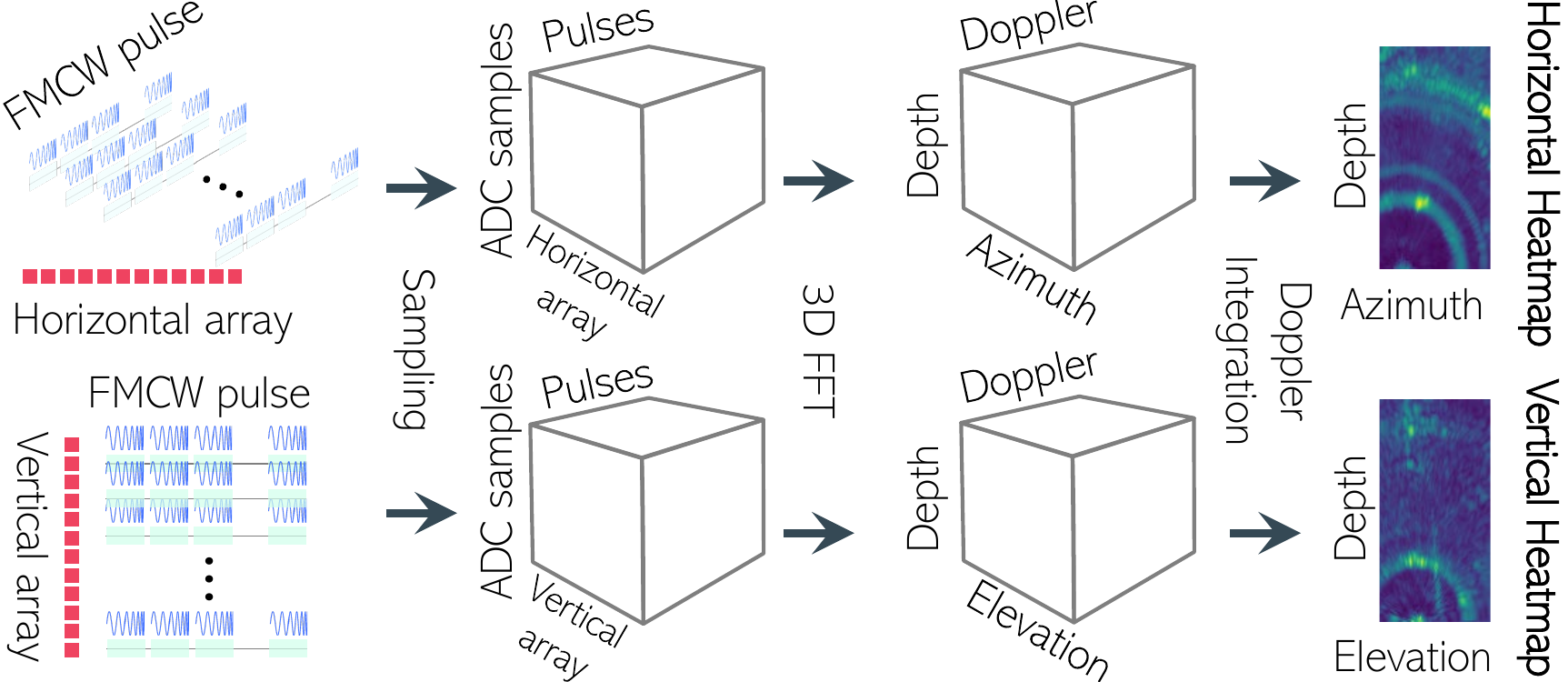}
    \caption{Multi-view heatmap preprocessing.}
    \label{fig:radar}
\end{wrapfigure} 
Fig.~\ref{fig:radar} illustrates the preprocessing flow of the multi-view radar heatmap using data from two MIMO-FMCW radars, which create two orthogonal virtual arrays composed of $86$ elements spaced at half-wavelength intervals while transmitting multiple pulses. By sampling the pulses reflected back, a 3D data cube can be formed, which is structured along the horizontal/vertical arrays, ADC samples (intra-pulse or fast-time), and pulse samples (inter-pulse or slow-time).
Performing a 3D fast Fourier transform (FFT) on this data cube yields radar spectra across the angle (azimuth for horizontal radar and elevation for vertical radar), range, and Doppler velocity domains. The SNR is further improved by integrating the 3D radar spectra along the Doppler domain, resulting in two radar heatmaps (range-azimuth and range-elevation) in polar radar coordinates. These heatmaps are then projected into the radar Cartesian coordinate system.

\section{Details of Experimental Settings}
\label{sec:experimental_setting}

\paragraph{MMVR Dataset:}
MMVR \cite{Perry2024_MMVR} has 345K data frames collected from $25$ human subjects over $6$ different rooms (e.g, open/cluttered offices and meeting rooms) spanning over $9$ separate days. MMVR consists of 2 parts: 1) 107.9K data frames of protocol 1 (P1): Open Foreground in a single open-foreground space with a single subject; and 2) 237.9K data frames of protocol 2 (P2): Cluttered space in 5 cluttered rooms with multiple subjects and multiple actions, including sitting postures. Data splits are set as same as S1 in MMVR. ``P1'' is used to establish the best possible radar perception benchmarks, while ``P2'' is designed for more challenging scenarios and for cross-environment and cross-subject generalization. The ``P2'' includes data such as sitting postures; therefore, we select this as the main dataset that we use in our experiments.

\paragraph{HIBER Dataset:}
\begin{wrapfigure}[10]{r}{2.6in}
    \centering
    \vspace{-0.15in}
    \includegraphics[width=0.47\textwidth]{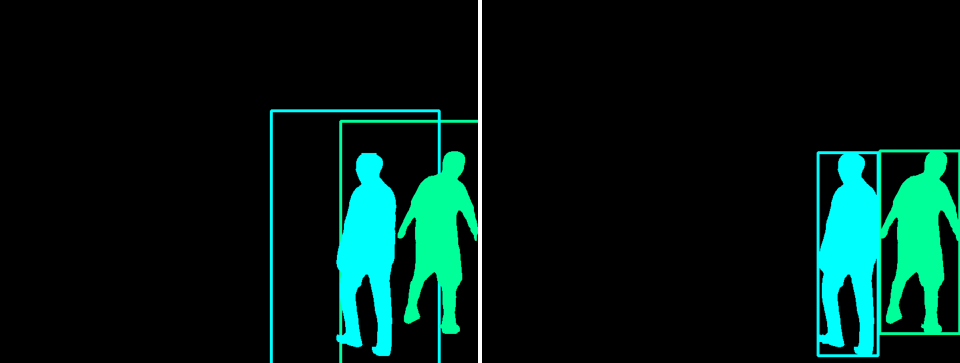}
    \caption{Original (left) versus Refined (right) BBoxes in the HIBER dataset. }
    \label{fig:hiber_refinement}
\end{wrapfigure}
HIBER~\cite{Wu2023_RFMask} is an open-source multi-view radar dataset including horizontal and vertical radar heatmaps and annotations such as 2D and 3D poses, BBoxes, and segmentation masks.
Among its data splits, ``WALK'' and ``MULTI'' are currently accessible. The “WALK” split includes 73.5K data frames, each featuring a single person per frame, while “MULTI” consistently includes two individuals per frame. 
We refined the original BBox labels in the HIBER dataset, addressing their initial overestimation by creating tighter BBoxes; see Fig.~\ref{fig:hiber_refinement} for an illustration.

\paragraph{Hyper-parameters:}
The hyper-parameters used in our experiments of Section~\ref{sec:experiments} are shown in Table~\ref{tab:detailed_hyperparameters}. The table is divided into three parts, Data, Model, and Training, each with parameter names, notations, and values for each dataset.

\begin{table}[t]
    \centering
    \footnotesize
    \caption{Details of hyper-parameters. Fixed-height size for HIBER dataset is depend on the environment.}
    \begin{tabular}{clcccc}
        \toprule
        \multicolumn{2}{c}{\multirow{2}{*}{\textbf{Name}}} & \multirow{2}{*}{\textbf{Notation}} & \multicolumn{2}{c}{\textbf{Value}} \\
        \cline{4-5}
         & & & \rule{0pt}{10pt}P2S1 / P2S2 & WALK / MULTI \\
        \midrule
        \multirow{9}{*}{\rotatebox{90}{\textbf{Data}}}
         & \# of training & - & 190441 / 118280 & 58382 / 53690 \\
         & \# of validation & - & 23899 / 33841 & 3229 / 8260 \\
         & \# of test & - & 23458 / 85677 & 11931 / 8850 \\
         & Input radar heatmap size & $H\times W$ & 256$\times$128 & 160$\times$200 \\
         & Segmentation mask size & $H\times W$ & 240$\times$320 & 624$\times$820 \\
         & Resolution of range & cm & 11.5 & 12.2\\
         & Resolution of azimuth & deg. &  1.3 &  1.3\\
         & Resolution of elevation & deg. &  1.3 &  1.3\\
         & Scale & - & log & - \\
        \midrule
        \multirow{14}{*}{\rotatebox{90}{\textbf{Model}}}
         & Backbone & - & ResNet18 & ResNet18 \\
         & Total dimension of positional embedding & - & 256 & 256  \\
         & Ratio of depth dimension for TPE & $\alpha$ & 0.6 & 0.6 \\
         & \# of input frames & - & 4 & 4 \\
         & Extracted feature map size & $H/s \times W/s$ & 64$\times$32 & 40$\times$56 \\
         & Top-$K$ selection & - & magnitude & magnitude \\
         & Top-$K$ & $K$ & 256 & 256 \\
         & \# of encoder blocks & $L_{\OP{self}}$ & 6 & 6 \\
         & \# of decoder blocks & $L_{\OP{cross}}$ & 6 & 6 \\
         & \# of head of multi-head attention & $M$ & 4 & 4 \\
         & \# of queries & $N$ & 10 & 10 \\
         & Threshold for detection and segmentation & - & 0.5 & 0.5 \\
         & Fixed-height size (pixel) & $H$ & 36 & - \\
         & Learning Transformation & - & True & False \\
        \midrule
        \multirow{20}{*}{\rotatebox{90}{\textbf{Training}}} 
         & Loss weight for $\OP{GIoU}$ on horizontal plane & $\lambda^{\OP{GIoU}}_{\OP{hor}}$ & 0.5 & 1.0 \\
         & Loss weight for $\OP{GIoU}$ on vertical plane & $\lambda^{\OP{GIoU}}_{\OP{ver}}$ & 0.5 & 1.0 \\
         & Loss weight for $\OP{GIoU}$ on image plane & $\lambda^{\OP{GIoU}}_{\OP{image}}$ & 1.0 & 1.0 \\
         & Loss weight for $\OP{L_1}$ on horizontal plane & $\lambda^{\OP{L_1}}_{\OP{hor}}$ & 0.5 & 1.0 \\
         & Loss weight for $\OP{L_1}$ on vertical plane & $\lambda^{\OP{L_1}}_{\OP{ver}}$ & 0.5 & 1.0 \\
         & Loss weight for $\OP{L_1}$ on image plane & $\lambda^{\OP{L_1}}_{\OP{image}}$ & 1.0 & 1.0 \\
         & Batch size & - & 32 & 32 \\
         & Epoch for detection & - & 100 & 100 \\
         & Epoch for segmentation & - & 20 & 20 \\
         & Patience for early stopping & - & 5 & 5 \\
         & Check val every $N$ epoch for early stopping  & - & 2 & 2 \\
         & Optimizer & - & AdamW & AdamW \\
         & Learning rate & - & 1e-4 & 1e-4 \\
         & Sheduler & - & Cosine & Cosine \\
         & Maximum number of epochs for sheduler & - & 100 & 100 \\
         & Weight decay & - & 1e-3 & 1e-3 \\
         & \# of workers & - & 8 & 8 \\
         & GPU (NVIDIA) & - & A40 & A40 \\
         & \# of GPUs & - & 1 & 1 \\
         & Approximate training time & day & 2 & 2 \\
        \bottomrule
    \end{tabular}
    \label{tab:detailed_hyperparameters}
\end{table}

\paragraph{RFMask with Refined BBoxes:}
We use RFMask~\cite{Wu2023_RFMask} as conventional method for BBox and segmentation tasks. 
However, RFMask can only predict relaxed BBoxes in the image plane due to its loss calculation being limited to the horizontal plane. Therefore, to train and predict using HIBER dataset (and also MMVR dataset), which consists of refined BBoxes as explained above, an additional module is required to convert the relaxed BBoxes predicted in the image plane into refined BBoxes.
As a result, we modify RFMask in a way that the BBox loss is calculated on the image plane and backpropagates to learnable parameters in an end-to-end fashion. 
Specifically, we add an image BBox regression module alongside a horizontal BBox Regression module, enabling the conversion of BBox offsets to the image plane. By computing loss with respect to these offsets, we can learn refined BBoxes on the image plane. Additionally, the region proposals estimated by the region proposal network (RPN) are transformed into 3D BBoxes based on the fixed-height size, the same as the original RFMask. These BBoxes are then projected onto the image plane and a 3D-to-2D projection. 

\section{Definition of Metrics}
\label{sec:definition_metrics}

\paragraph{Mean Intersection over Union:}
We adopt average precision on intersection over union~(IoU)~\cite{Everingham2010_pascalvoc} as an evaluation metric. 
IoU is the ratio of the overlap to the union of a predicted BBox $A$ and annotated BBox $B$ as:
\begin{equation}\label{eq:iou}
    \OP{IoU}\LS{A, B} = \frac{|A \bigcap B|}{|A \bigcup B|}.
\end{equation}

\paragraph{Average Precision:}
Average Precision (AP) can then be defined as the area under the interpolated precision-recall curve, which can be calculated using the following formula:
\begin{align}
    \OP{AP} &= \sum_{i=1}^{n-1}\LS{r_{i+1}-r_i} p_{\OP{interp}}\LS{r_{i+1}}\\
    p_{\OP{interp}}\LS{r} &= \max _{r^{\prime} \geq r} p\LS{r^{\prime}},
\end{align}
where The interpolated precision 
$p_{\OP{interp}}$ at a certain recall level $r$ is defined as the highest precision found for any recall level $r^{\prime} \geq r$.
We present three variants of average precision: $\OP{AP}_{50}$, $\OP{AP}_{75}$, and $\OP{AP}$, where the former two represent the loose and strict constraints of IoU, while $\OP{AP}$ is the averaged score over $10$ different IoU thresholds in $[0.5, 0.95]$ with a stepsize of $0.05$.

\paragraph{Average Recall:}
Average recall (AR)~\cite{Hosang2016_AverageRecall} between 0.5 and 1 of $\OP{IoU\; overlap\; threshold}$ can be computed by averaging over the overlaps of each annotation $\mathrm{gt}_{i}$ with the closest matched proposal, that is integrating over the $y:\OP{recall}$ axis of the plot instead of the $x:\OP{IoU\; overlap\; threshold}$ axis. Let $o$ be the $\OP{IoU}$ overlap and $\OP{recall}\LS{o}$ the function. Let $\OP{IoU}\LS{\mathrm{gt}_{i}}$ denote the $\OP{IoU}$ between the annotation $\mathrm{gt}_{i}$ and the closest detection proposal:
\begin{equation}
\OP{AR}=2 \int_{0.5}^1 \OP{recall}(o) \mathrm{d} o=\frac{2}{n} \sum_{i=1}^n \max \LS{\OP{IoU}\LS{\mathrm{gt}_i}-0.5,0}.
\end{equation}
The followings are some variations of $\OP{AR}$:
\begin{itemize}
    \item $\OP{AR}_{1}$: $\OP{AR}$ given 1 detection per data.
    \item $\OP{AR}_{10}$: $\OP{AR}$ given 10 detection per data.
    \item $\OP{AR}_{100}$: $\OP{AR}$ given 100 detection per data.
\end{itemize}

\section{Additional Ablation Study}
\label{sec:additional_ablation_study}

To validate the effectiveness of our RETR, we conducted additional ablation studies. Unless otherwise specified, the hyperparameters follow those listed in the Table~\ref{tab:detailed_hyperparameters}.

\paragraph{Results under ``MULTI'' on HIBER dataset:}
Table~\ref{tab:main_hiber_multi} shows the evaluation results on the HIBER dataset. From this table, it can be seen that using RETR improves performance across all metrics. Similar to the MMVR results, RETR with tri-plane loss shows enhanced performance. Additionally, the use of RETR with TPE in both the encoder and decoder also contributes to performance improvements. However, compared to the ``P2S1'' of MMVR, the performance improvement is smaller (the improvement is 15.28 $\OP{AP}$ from DETR to RETR). This is likely because, unlike ``P2S1'', HIBER under ``WALK'' only involves walking actions, which benefit less from the use of 3D information.

\begin{table}[t]
    \footnotesize
    \centering
    \caption{Results under ``MULTI'' on HIBER dataset.}
    \begin{tabular}{lccc| ccc cccc}
        \toprule
        \multicolumn{1}{c}{$\OP{Model}$} & $\OP{Dim}$ & $\OP{Input}$ & $\OP{BBox} \ \OP{Loss}$ & $\OP{AP}$ & $\OP{AP_{50}}$ & $\OP{AP_{75}}$ & $\OP{AR_{1}}$ & $\OP{AR_{10}}$ \\
        \midrule
        DETR & 2D & $\OP{H}$ & $\OP{H}+\OP{I}$ & 17.00 & 52.79 & 6.08 & 16.21 & 31.57  \\
        DETR (Top-$K$) & 2D & $\OP{H}$, $\OP{V}$ & $\OP{H}+\OP{I}$ & 22.74 & 60.64 & 11.66 & 18.62 & 37.62 \\
        \midrule
        RETR (TPE@Dec.) & 2D & $\OP{H}$, $\OP{V}$ & $\OP{H}+\OP{I}$ & 23.02 & 61.77 & 12.51 & 19.17 & 38.00 \\
        RETR & 2D & $\OP{H}$, $\OP{V}$ & $\OP{H}+\OP{I}$        & 23.53 & 63.84 & 11.90 & 20.37 & 38.16 \\
        \midrule
        RETR (TPE@Dec.) & 3D & $\OP{H}$, $\OP{V}$ & $\OP{H}+\OP{V}+\OP{I}$ & 26.36 & 69.50 & 14.33 & 20.76 & 40.90 \\
        RETR & 3D & $\OP{H}$, $\OP{V}$ & $\OP{H}+\OP{V}+\OP{I}$ & \cellcolor{gray!20}28.98 & \cellcolor{gray!20}74.82 & \cellcolor{gray!20}15.64 & \cellcolor{gray!20}22.95 & \cellcolor{gray!20}41.18 \\
        \bottomrule
    \end{tabular}
    \label{tab:main_hiber_multi}
\end{table}

\begin{table}[t!]
    \footnotesize
    \centering
    \caption{Results under ``P2S2'' on MMVR dataset.}
    \begin{tabular}{lccc| ccc cccc}
        \toprule
        \multicolumn{1}{c}{$\OP{Model}$} & $\OP{Dim}$ & $\OP{Input}$ & $\OP{BBox} \ \OP{Loss}$ & $\OP{AP}$ & $\OP{AP_{50}}$ & $\OP{AP_{75}}$ & $\OP{AR_{1}}$ & $\OP{AR_{10}}$ \\
        \midrule
        RFMask & 2D & $\OP{H}$, $\OP{V}$ & $\OP{H}+\OP{I}$ & 6.03 & 22.77 & 0.88 & 9.25 & 12.09  \\
        DETR (Top-$K$) & 2D & $\OP{H}$, $\OP{V}$ & $\OP{H}+\OP{I}$ & 9.29 & 34.69 & 2.49 & \cellcolor{gray!20}20.68 & \cellcolor{gray!20}22.82 \\
        \midrule
        RETR & 2D & $\OP{H}$, $\OP{V}$ & $\OP{H}+\OP{I}$  & 10.37 & 35.40 & 3.36 & 18.83 & 20.06 \\
        RETR & 3D & $\OP{H}$, $\OP{V}$ & $\OP{H}+\OP{V}+\OP{I}$ & \cellcolor{gray!20}12.19 & \cellcolor{gray!20}40.67 & \cellcolor{gray!20}4.95 & 19.70 & 21.34 \\
        \bottomrule
    \end{tabular}
    \label{tab:main_mmvr_p2s2}
\end{table}

\begin{table}[t!]
        \footnotesize
        \centering
        \caption{Impact of \textbf{Learnable Transformation} ($\OP{LT}$) with additional precision and recall metrics. }
        \begin{tabular}{c ccc cccc}
            \toprule
            $\OP{LT}$ & $\OP{AP}$ & $\OP{AP_{50}}$ & $\OP{AP_{75}}$ & $\OP{AR_{1}}$ & $\OP{AR_{10}}$ \\
            \midrule
            -           & 42.97 & 80.54 & 41.97 & 40.20 & 55.58 \\
            \checkmark  & \cellcolor{gray!20}46.75 & \cellcolor{gray!20}83.80 & \cellcolor{gray!20}46.06 & \cellcolor{gray!20}42.19 & \cellcolor{gray!20}57.39 \\
            \bottomrule
        \end{tabular}
        \label{tab:abl_mmvr_learnable_transformation_full}
\end{table}

\begin{table}[t]
    \footnotesize
    \centering
    \caption{Full table: \textbf{Tri-plane loss} can improve the performance.}
    \begin{tabular}{cccccc c}
        \toprule
        $\OP{BBox} \ \OP{Loss}$ & $\OP{AP}$ & $\OP{AP_{50}}$ & $\OP{AP_{75}}$ & $\OP{AR_{1}}$ & $\OP{AR_{10}}$ & $\OP{IoU}$ \\
        \midrule
        $\OP{H}+\OP{I}$ & 42.18 & 80.49 & 39.24 & 39.39 & 53.84 & 75.15 \\
        $\OP{H}+\OP{V}+\OP{I}$ & \cellcolor{gray!20}46.75 & \cellcolor{gray!20}83.80 & \cellcolor{gray!20}46.06 & \cellcolor{gray!20}42.19 & \cellcolor{gray!20}57.39 & \cellcolor{gray!20}77.21 \\
        \bottomrule
    \end{tabular}
    \label{tab:abl_3d_prediction_full}
\end{table}

\begin{table}[t!]
    \footnotesize
    \centering
    \caption{Impact of the value of $K$ on object detection}
        \begin{tabular}{c ccc cccc}
            \toprule
            $K$ & $\OP{AP}$ & $\OP{AP_{50}}$ & $\OP{AP_{75}}$ & $\OP{AR_{1}}$ & $\OP{AR_{10}}$ \\
            \midrule
            64  & 38.39 & 80.00 & 31.85 & 37.95 & 50.79 \\
            100 & 38.82 & 80.06 & 32.99 & 38.85 & 52.28 \\
            196 & \cellcolor{gray!20}46.97 & 82.65 & 45.55 & \cellcolor{gray!20}42.93 & \cellcolor{gray!20}57.62 \\
            256 & 46.75 & \cellcolor{gray!20}83.80 & \cellcolor{gray!20}46.06 & 42.19 & 57.39 \\
            \bottomrule
        \end{tabular}
    \label{abl_mmvr_topk}
\end{table}

\begin{table}[t!]
    \footnotesize
    \centering
    \caption{Impact of training data size on object detection}
        \begin{tabular}{c ccc cccc}
            \toprule
            \# of data & $\OP{AP}$ & $\OP{AP_{50}}$ & $\OP{AP_{75}}$ & $\OP{AR_{1}}$ & $\OP{AR_{10}}$ \\
            \midrule
            $\times 0.1$  & 37.10 & 78.30 & 31.30 & 37.50 & 50.33 \\
            $\times 0.5$  & 40.84 & 79.80 & 36.82 & 40.06 & 53.33 \\
            $\times 1.0$  & \cellcolor{gray!20}46.75 & \cellcolor{gray!20}83.80 & \cellcolor{gray!20}46.06 & \cellcolor{gray!20}42.19 & \cellcolor{gray!20}57.39 \\
            \bottomrule
        \end{tabular}
    \label{tab:abl_mmvr_num_training}
\end{table}

\begin{table}[t!]
    \centering
    \footnotesize
    \caption{\textbf{Inference time} and frame rate (FPS) of {RFMask and RETR}.}
    \begin{tabular}{ccc}
        \toprule
        Method & Time[ms] & FPS \\
        \midrule
        RFMask  & \cellcolor{gray!20}20.89 & \cellcolor{gray!20}47.87\\
        RETR    & 23.75 & 42.11 \\
        \bottomrule
    \end{tabular}
    \label{tab:abl_mmvr_inference_time}
\end{table}

\paragraph{Results under ``P2S2'' on MMVR dataset:}
``P2S2'' (Cross-Session and Unseen Split) on the MMVR dataset first splits all data segments in d5, d6, d7, and d9 into train, validation, and test sets. Then, it is included all data in d8 in the test set such that one can assess the generalization performance of trained model for an unseen environment (d8). Therefore, ``P2S2'' is the most challenging scenario in the MMVR.
Table~\ref{tab:main_mmvr_p2s2} shows the evaluation results under ``P2S2''. 
From the results in the table, we confirmed that the prediction performance was improved by using RETR. In particular, RETR outperforms RFMask by a margin of $11.92+ \OP{AP}_{50}$.
However, compared to the results of P2S1, there is a significant decrease in performance, and this is due to the unseen environment.

\paragraph{Impact of Tunable Dimension Ratio $\alpha$ in TPE:}
\begin{wrapfigure}[17]{r}{2.7in}
    \vspace{-0.02in}
    \centering
    \includegraphics[width=0.49\textwidth]{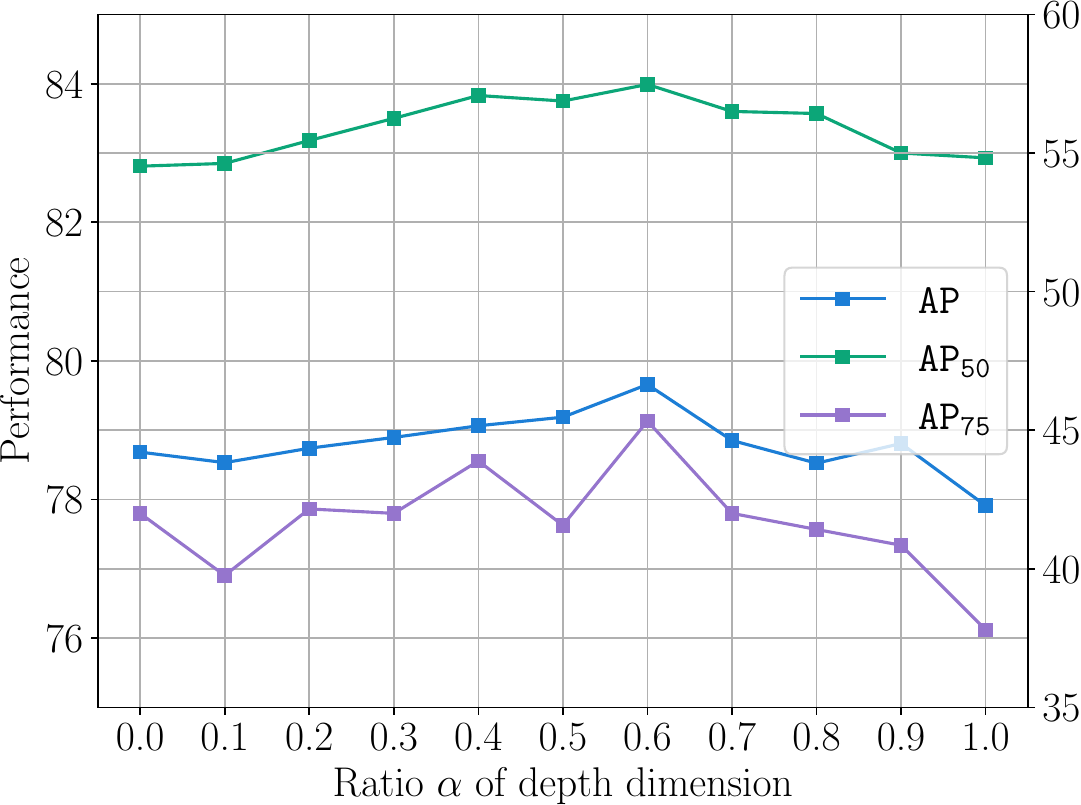}
    \caption{Full figure: Impact of \textbf{tunable dimension ratio $\alpha$ in TPE}.}
    \label{fig:ex_abl_ratio_full}
\end{wrapfigure}
To investigate the impact of tuning in TPE, we observed the performance differences when varying the ratio of depth and angle dimensions (since the total dimension is $d_{\OP{pos}} = 256$, e.g., a ratio $\alpha=0.2$  means depth is rounded to $d_{\OP{dep}} =\alpha d_{\OP{pos}} = 102$ dimensions and angle to $d_{\OP{ang}} =(1-\alpha) d_{\OP{pos}} = 154$ dimensions). In Section~\ref{sec:ablation_study}, we showed detection results in Table~\ref{tab:abl_mmvr_learnable_transformation} with selected some $\alpha$ and metrics. Here, we show the results with more variants of $\alpha$ and metrics.
Fig.~\ref{fig:ex_abl_ratio_full} shows the result. The horizontal axis denotes the proportion of depth dimensions, and the vertical axis denotes the performance of various metrics. Note that $\OP{AP_{50}}$ refers to the primary axis, while $\OP{AP}$ and $\OP{AP_{75}}$ refer to the secondary axis. The figure shows that the highest performance is achieved when the depth proportion is $\alpha = 0.6$. The performance exhibits a peak at this point, indicating that prioritizing depth improves performance.

\paragraph{Impact of Learnable Transformation:}
We expand Table~\ref{tab:abl_mmvr_learnable_transformation} by including additional precision and recall metrics, providing a more comprehensive evaluation of the Learnable Transformation. The complete results are presented in Table~\ref{tab:abl_mmvr_learnable_transformation_full}. 

\paragraph{Tri-Plane Loss:}
In Section~\ref{sec:learnTrans}, Table~\ref{tab:abl_3d_prediction} compared RETR with a bi-plane BBox loss (horizontal radar plane and image plane) to that with the tri-plane loss (including the vertical radar plane).  
We show the more results with complete metrics. The results in Table~\ref{tab:abl_3d_prediction_full} highlight the necessity of accounting for the vertical BBox loss and the importance of leveraging features from the vertical radar heatmap.

\paragraph{Impact of the Value of $K$ in Top-$K$ Selection:}
Table~\ref{abl_mmvr_topk} shows that, as the value of $K$ increases (e.g., $K=196$ and $K=256$),  object detection performance improves across both precision and recall metrics. 

\paragraph{Impact of Training Data Size:}
Table~\ref{tab:abl_mmvr_num_training} reports the effect of training data size on detection performance using the MMVR dataset. We compare the original data size ($\times 1.0$) with $190,441$ radar frames against reduced data sizes of half ($\times 0.5$) and one-tenth ($\times 0.1$).  The results demonstrate a gradual improvement in detection performance as the data size increases.

\paragraph{Inference Time:}
Table~\ref{tab:abl_mmvr_inference_time} reports the average inference time in milliseconds, evaluated over all frames in the test data using an NVIDIA A40 GPU.  RETR achieves an average inference time of $23.75$ ms, which is comparable to that of RFMask at $20.89$ ms. 

\section{Visualization Result}
\label{sec:visualization_result}

\paragraph{Comparison:}
\begin{figure*}[t]
    \centering
    \includegraphics[width=1.0\textwidth]{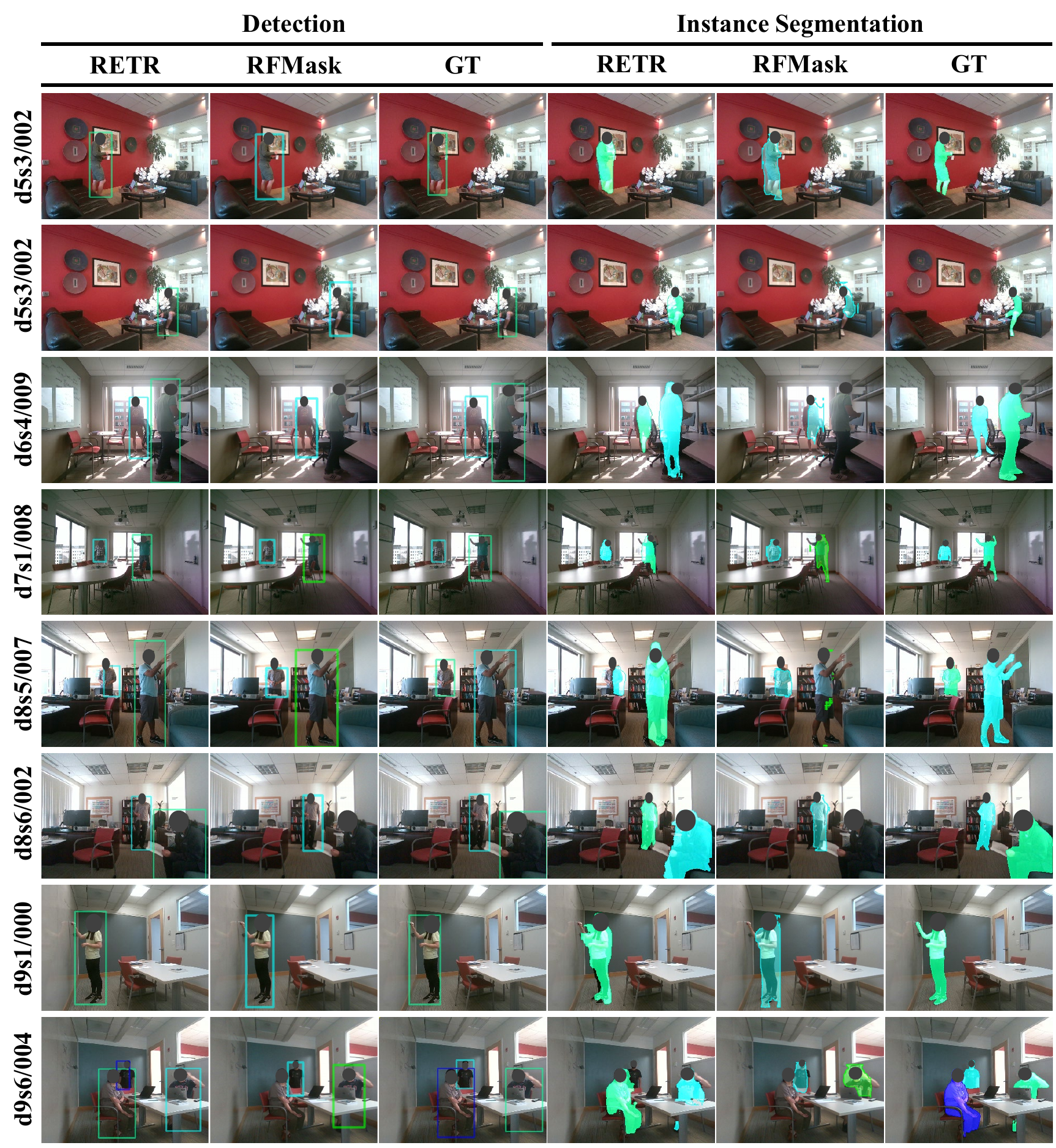}
    \caption{Visualization and comparison between RETR and RFMask. Each row indicates the segment name used from the ``P2S1'' test dataset.} 
    \label{fig:vis_results_for_detection}
\end{figure*}

To compare the conventional RFMask with our RETR, we visualized the prediction results of each. Fig.~\ref{fig:vis_results_for_detection} shows these results. Each row indicates the segment name used from the ``P2S1'' test dataset. In detection, RFMask has some miss-detections, whereas RETR accurately predicts even when there are multiple subjects. For instance, RFMask tends to fail in detecting people close to the camera, as seen in d8s6/002, but this is improved with RETR. In segmentation, RETR captures human shapes more accurately than RFMask. However, RETR can also fail in mask estimation, as seen in the example of d9s6.

\paragraph{Analysis of Failure Cases:}
\begin{figure*}[t]
    \centering
    \includegraphics[width=0.75\textwidth]{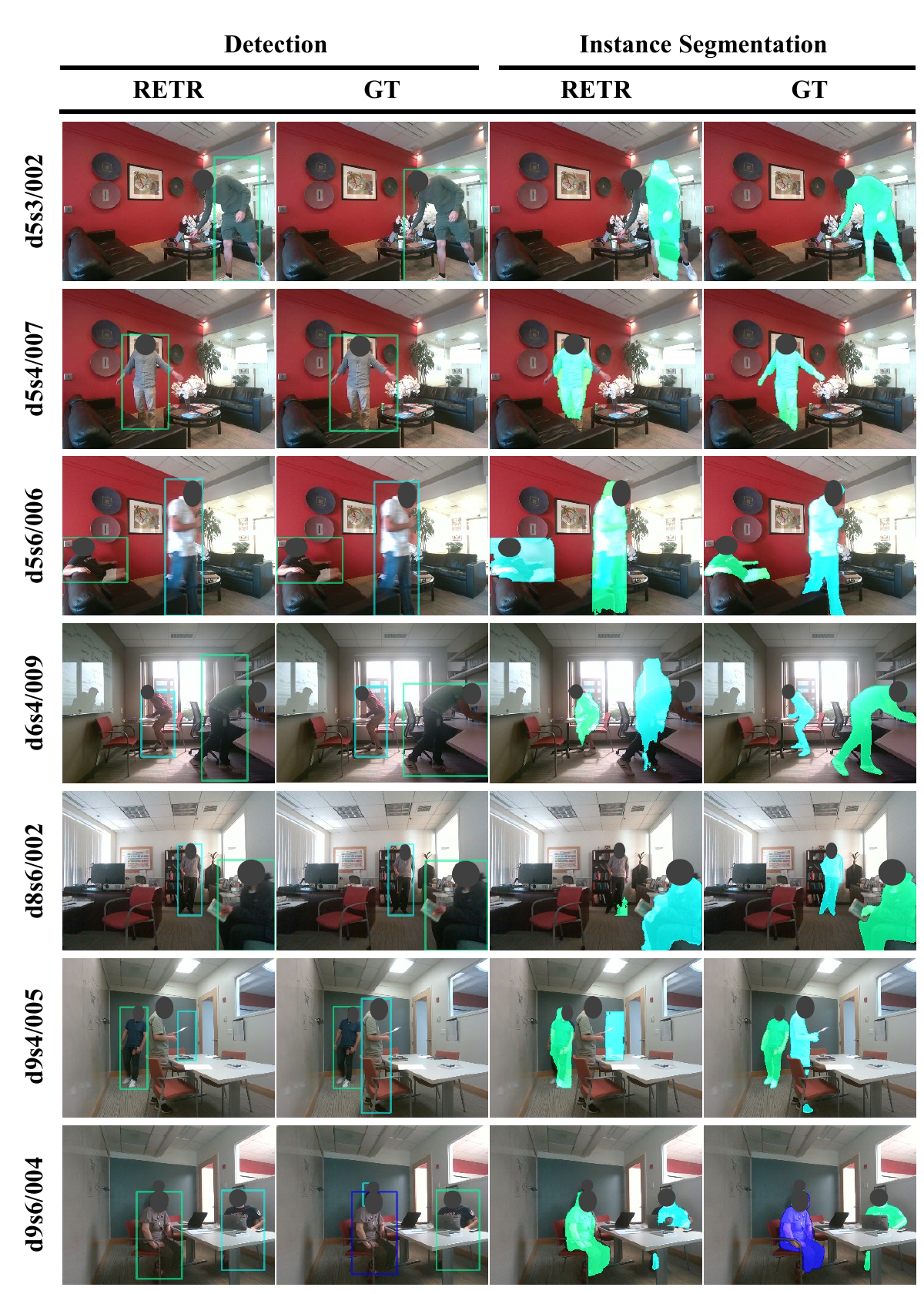}
    \caption{Visualization of failure cases. Each row indicates the segment name used from the ``P2S1'' test dataset.} 
    \label{fig:vis_results_for_detection_fail}
\end{figure*}

We provide failure cases in Fig.~\ref{fig:vis_results_for_detection_fail}. As shown in images such as d5s3/002 and d6s4/009, RETR occasionally mispredicts a bending-over person as standing. Additionally, as shown in d5s4/007, it is often challenging to predict the detailed position of the arms, leading to failures in both detection and segmentation. In some cases, such as d5s6/006 and d8s6/002, the segmentation mask region was excessively large or too narrow. Moreover, in instances such as d9s6/004, while the BBox prediction was successful, the segmentation failed. There was also a case, such as d9s4/005, where the inaccuracy in the BBox prediction led to an incorrect mask position.

\end{document}